\documentclass{article}
\usepackage{booktabs}
\usepackage{multirow}
\usepackage{graphicx}
\usepackage{amsmath}  
\usepackage{microtype}
\usepackage{graphicx}
\usepackage{subcaption}
\usepackage{booktabs} 
\usepackage{tabularx}
\usepackage{multicol}

\usepackage{hyperref}

\usepackage[preprint]{icml2026}

\usepackage{amsmath}
\usepackage{amssymb}
\usepackage{mathtools}
\usepackage{amsthm}

\usepackage[utf8]{inputenc}
\usepackage{multirow}   
\usepackage{amssymb}    
\usepackage{enumitem}

\usepackage[capitalize,noabbrev]{cleveref}

\theoremstyle{plain}

\theoremstyle{definition}

\theoremstyle{remark}

\usepackage[textsize=tiny]{todonotes}

\begin{document}

\twocolumn[
  \icmltitle{HII-DPO: Eliminate Hallucination via Accurate Hallucination-Inducing Counterfactual Images}



  
  \icmlsetsymbol{equal}{*}
  \icmlsetsymbol{corresponding}{\ensuremath{\dagger}}

  \begin{icmlauthorlist}
    \icmlauthor{Yilin Yang}{uh}
    \icmlauthor{Zhenghui Guo}{uh}
    \icmlauthor{Yuke Wang}{rice}
    \icmlauthor{Omprakash Gnawali}{uh}
    \icmlauthor{Sheng Di}{ANL}
    \icmlauthor{Chengming Zhang}{uh,corresponding}

  \end{icmlauthorlist}

  \icmlaffiliation{uh}{University of Houston}
  \icmlaffiliation{rice}{Rice University}
  \icmlaffiliation{ANL}{Argonne National Laboratory}

  \icmlcorrespondingauthor{Chengming Zhang}{czhang59@central.uh.edu}

  \icmlkeywords{Machine Learning, ICML}
  \vskip 0.3in
]



\printAffiliationsAndNotice{}  

\begin{abstract}


Large Vision-Language Models (VLMs) have achieved remarkable success across diverse multimodal tasks but remain vulnerable to hallucinations rooted in inherent language bias. Despite recent progress, existing hallucination mitigation methods often overlook the underlying hallucination patterns driven by language bias. In this work, we design a novel pipeline to accurately synthesize Hallucination-Inducing Images (HIIs). Using synthesized HIIs, we reveal a consistent scene-conditioned hallucination pattern: models tend to mention objects that are highly typical of the scene even when visual evidence is removed. To quantify the susceptibility of VLMs to this hallucination pattern, we establish the Masked-Object-Hallucination (MOH) benchmark to rigorously evaluate existing state-of-the-art alignment frameworks. Finally, we leverage HIIs to construct high-quality preference datasets for fine-grained alignment. Experimental results demonstrate that our approach effectively mitigates hallucinations while preserving general model capabilities. Specifically, our method achieves up to a 38\% improvement over the current state-of-the-art on standard hallucination benchmarks.

\end{abstract}

\section{Introduction}
\label{sec::intro}

Large Vision-Language Models (VLMs) demonstrate remarkable capabilities in jointly perceiving and reasoning over images and text, facilitating applications in assistive technologies, education, medical imaging, and industrial inspection. Despite this outstanding progress, modern VLMs are still prone to hallucinations, producing fluent yet visually incorrect content or misaligning visual evidence with textual outputs \cite{rohrbach2018object,zhao2023evaluating,xiong2024survey}. Such inconsistency drastically undermines the trustworthiness of AI systems and hinders the widespread applications of VLMs.

Recent research~\cite{wang2024mdpoconditionalpreferenceoptimization, xing2025realignaligningvisionlanguage, chen2025perturbollavareducingmultimodalhallucinations, zadeh2025lpoilistwisepreferenceoptimization, zhou2024analyzingmitigatingobjecthallucination, zhou2024aligningmodalitiesvisionlarge} indicates that VLMs are prone to modality bias, a phenomenon where generation is disproportionately guided by linguistic priors at the expense of visual evidence. For example, a VLM may instinctively generate the term “sink” following the mention of “kitchen”, or a “keyboard” following the mention of an “office”, relying on textual co-occurrences learned during training. Consequently, visual details are overlooked, resulting in severe hallucinations and compromised model trustworthiness.  This issue stems primarily from an architectural imbalance in current VLMs, where the Large Language Model (LLM) component dominates the visual encoder. Consequently, VLMs inevitably rely on the linguistic priors, rather than grounding their responses in visual evidence. This phenomenon is referred to as \textbf{Language Bias} or \textbf{Language Prior}.

These deficiencies have motivated a shift toward Direct Preference Optimization (DPO)~\cite{rafailov2024directpreferenceoptimizationlanguage,wang2024mdpoconditionalpreferenceoptimization,xing2025realignaligningvisionlanguage,zhou2024aligningmodalitiesvisionlarge,zhang2025mmrlhfstepforwardmultimodal} which is emerging as a crucial approach to mitigate language bias and enhance visual faithfulness. Specifically, mDPO~\cite{wang2024mdpoconditionalpreferenceoptimization} pioneered the paradigm of utilizing contrastive visual signals for preference optimization---a framework we define as Vision-Contrastive Alignment (VCA). VCA-based methods compel VLMs to prioritize visual modalities by leveraging visual contrastive preferences. However, current VCA-based approaches are hindered by three critical limitations. First, these implementations suffer from semantic ambiguity. The chosen responses often retain semantic details that remain partially grounded in the rejected visual inputs, resulting in a coarse-grained alignment signal that lacks the resolution to distinguish specific object-level absences. Second, the construction of rejected images via stochastic processes, such as random cropping or similarity-based retrieval, is fundamentally insufficient to generate “hard” negative samples, as it often fails to preserve the scene context necessary to trigger linguistic priors. Third, existing VCA-based frameworks fail to leverage negative images to construct chosen and rejected responses for preference alignment. Consequently, VLMs do not learn a deeper understanding of the negative images, leading to suboptimal performance on various benchmarks. Therefore, it remains challenging to generate precise, “hard” counterfactual samples or preference datasets that can effectively decouple linguistic priors from visual evidence during preference alignment.


\vspace*{-10px}


\paragraph{Our Observation.}

We revisit object hallucination from the perspective of \textbf{scene-object co-occurrence}, i.e., the tendency for certain objects to frequently appear in particular scenes. Such co-occurrence bias can shape the model's language prior in DDG. For instance, VLMs often mention \emph{“sink”} in the kitchen, \emph{“stop sign”} at crossroads, or \emph{“train”} on railways, as these objects are highly probable given the surrounding context. To investigate this effect, we construct counterfactual images by deliberately removing salient objects from the input and then prompt VLMs to perform DDG. We find that VLMs across a range of model scales and architectures still \emph{falsely assert the presence} of the removed objects even when there is no supporting visual evidence. We refer to this failure mode as \textbf{scene-conditioned hallucination}: hallucinations conditioned on the remaining scene context (in the image or the textual prompt) after object removed in the image, in which a model's predictions are {strongly influenced by} scene-object co-occurrence priors and thus can override grounded visual cues {under counterfactual inputs}. This phenomenon {may be partially explained by} the high frequency of co-occurrence between specific {scenarios} and objects in SFT corpora, which can reinforce such priors. However, existing approaches have not provided an effective way to quantitatively evaluate hallucination conditioned by the scene, or a principled strategy to mitigate it.

\vspace*{-10px}
\paragraph{Our Contributions.}

To investigate and quantify the scene-conditioned hallucination pattern, we introduce a systematic pipeline to synthesize accurate Hallucination-Inducing Images (HIIs). By strategically removing salient objects from natural images, we curate a set of high-quality, accurate counterfactual samples, where models are highly prone to hallucinating missing entities due to rooted linguistic bias. Utilizing these counterfactual images, we construct the Masked-Object-Hallucination (MOH) benchmark. This novel evaluation framework assesses a model's resilience when confronted with counterfactual images. This helps us to quantitatively analyze how existing state-of-the-art (SOTA) alignment frameworks mitigate this specific hallucination pattern. Finally, we propose HII-DPO, a novel data construction and preference alignment pipeline, designed to decouple language bias from visual grounding. Utilizing HIIs, HII-DPO synthesizes fine-grained preference pairs that explicitly guide the Direct Preference Optimization (DPO) process, effectively teaching models to remain faithful to visual inputs rather than relying on stereotypical scene-object co-occurrences.

We evaluate current SOTA methods on our newly proposed MOH benchmark, providing a rigorous quantitative analysis of hallucination rates under counterfactual scenarios. We also conduct extensive experiments on multiple benchmarks using HII-DPO. Specifically, our method achieves up to a 38\% improvement over existing SOTA methods and reduces the hallucination rate by up to 92\% compared to baseline models on the standard hallucination benchmarks~\cite{rohrbach2018object,wang2024amberllmfreemultidimensionalbenchmark}. Crucially, this gain does not come at the cost of general capabilities, as our model maintains robust performance on standard VQA tasks~\cite{Goyal_2017_CVPR,Singh_2019_CVPR,lu2022learn,yu2023mm}.

In summary, our contributions are four-fold:
\begin{itemize} 
\vspace*{-8px}
\item We devise a systematic pipeline to synthesize accurate HIIs. These counterfactual images effectively induce scene-conditioned hallucinations, providing a high-quality diagnostic tool to expose the linguistic bias rooted in VLMs.
\vspace*{-5px}
\item To quantitatively evaluate the scene-conditioned hallucination pattern, we curate the MOH benchmark. By rigorously testing existing approaches, we empirically demonstrate the prevalence of scene-object stereotypical predictions and provide a quantitative analysis of VLMs’ susceptibility to scene-conditioned hallucination.
\vspace*{-5px}
\item Leveraging HIIs, we propose a novel data construction and preference alignment framework, HII-DPO. Utilizing fine-grained preference pairs with shared prefixes, our pipeline effectively mitigates scene-conditioned hallucinations by teaching models to prioritize visual grounding over biased linguistic heuristics.
\vspace*{0px}
\item Extensive experimental results confirm that our approach establishes a new state-of-the-art across multiple hallucination benchmarks while preserving general multimodal capabilities. 
\end{itemize}

\section{Preliminaries}
\label{sec::preliminary}

In this section, we will briefly introduce the background of VLMs and direct preference optimization, serving as the theoretical basis for our approach.

\subsection{Large Vision Language Model}
Large Vision-Language Models (VLMs) aim to unify visual and linguistic understanding by extending large language models (LLMs) with visual perception modules. Typically, a VLM consists of three major components: a \textit{vision encoder}, a \textit{multimodal projector}, and a \textit{language decoder}. The vision encoder (e.g., CLIP~\cite{radford2021clip}, ViT~\cite{dosovitskiy2021vit}, or SigLIP~\cite{zhai2023siglip}) extracts dense visual representations from the input image. The multimodal projector aligns the visual embeddings with the token embedding space of the language model, enabling cross-modal interactions. Finally, the language decoder (e.g., LLaMA~\cite{touvron2023llama}, Vicuna~\cite{chiang2023vicuna}, or Qwen2~\cite{yang2024qwen2}) autoregressively generates textual outputs conditioned on both the visual context and the input instruction.

\noindent
\subsection{Direct Preference Optimization}
In order to improve performance of VLMs, Direct Preference Optimization (DPO)~\cite{rafailov2023direct} and its diverse variants~\cite{wang2024mdpoconditionalpreferenceoptimization,xing2025realignaligningvisionlanguage,zhou2024aligningmodalitiesvisionlarge,zhang2025mmrlhfstepforwardmultimodal} have been shown to be an efficient fine-tuning method that requires minimal additional computational resources. For the standard DPO in a multimodal setting, a preference pair is defined as $\mathcal{D} = \{ x,v,y^+,y^- \}$, where $y^-$ and $y^+$ represent preferred and dispreferred responses, respectively. $x$ is the input prompt and $v$ is the visual signal input. DPO is based on Bradley-Terry Model~(BT)~\cite{bradley1952rank}. The objective of the BT model is to maximize the likelihood of the observed preference pairs, i.e., the probability that the preferred response is ranked higher than the dispreferred one:
\begin{equation}
    p(y^+\succ y^-)=\sigma(r(x,v,y^+)-r(x,v,y^-))
    \label{eq:bt_model}
\end{equation}
In the DPO scenario, $r(x,v,y)$ is defined as the implicit reward of a specific preference data tuple $(x,v,y)$. For a single preference tuple $(x,v,y)$:
\begin{equation}
    r(x,v,y) = \beta \cdot \mathrm{log}\frac{ \ {\pi_\theta}{\left( y|x,v \right)}}{ \ {\pi_\mathrm{ref}}{\left( y|x,v \right)}}
    \label{eq:rw}
\end{equation}
Here $\pi_\mathrm{ref}$ is denoted as the reference model. $\pi_{\theta}$ is the policy model. Finally,~\autoref{eq:bt_model} can be reformulated in the form of a maximum-likelihood objective computed over the entire training dataset:
\begin{equation}
    \scalebox{0.92}{$
    \mathcal{L}_\mathrm{DPO}=-\log\sigma\left(\beta\log\frac{\pi_\theta(y^+|x,v)}{\pi_{\mathrm{ref}}(y^+|x,v)}-\beta\log\frac{\pi_\theta(y^-|x,v)}{\pi_{\mathrm{ref}}(y^-|x,v)}\right).
    $}
\end{equation}
Recent work \cite{xing2025realignaligningvisionlanguage,peng2025mitigatingobjecthallucinationssentencelevel,li2024hsadpo,he2024topic,yu2025rlaif} focuses on generating high-quality preference data for DPO to achieve the proper alignment between vision and language. 


\section{Method}

\begin{figure*}[htbp]
    \centering
    \resizebox{0.87\textwidth}{0.87\height}{\includegraphics[trim=0em 0em 0em 0em,clip,width=1\textwidth]{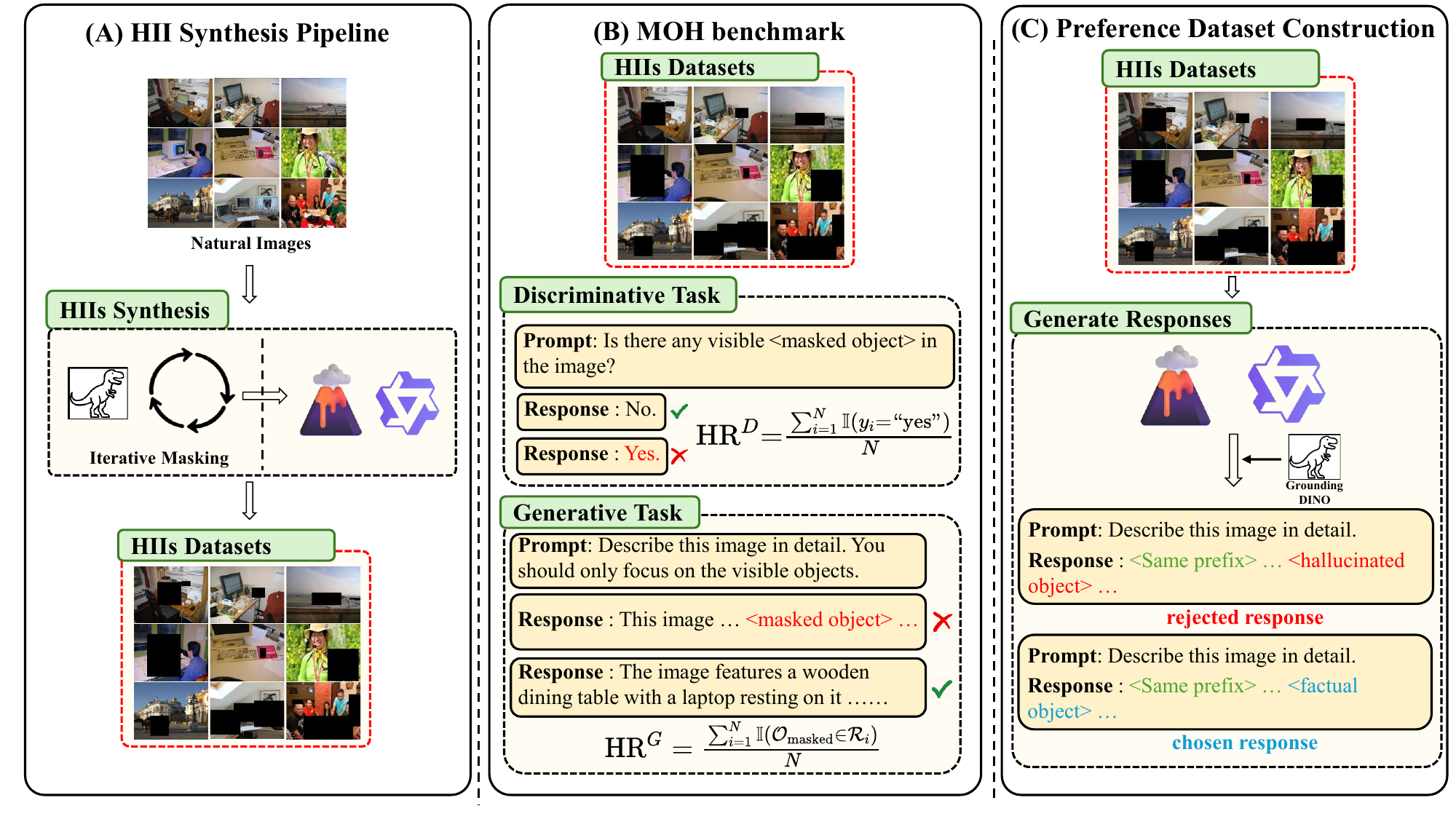} }
    \caption{\textbf{Overview of our framework.}~\textbf{(A)} Synthesize HIIs using GroundingDINO and open-source VLMs.~\textbf{(B)} Construct MOH benchmark to quantitatively evaluate the scene-conditioned hallucination pattern. \textbf{(C)} Generate preference dataset using HIIs.}
    \label{fig::overview}
\end{figure*}


\subsection{Overview}

We present the overview of our framework in~\autoref{fig::overview}. First, we propose a new pipeline for synthesizing HIIs based on our observations about language priors. Second, we design a new benchmark based on the counterfactual images to evaluate whether models can overcome language priors and scene-conditioned biases to make correct object-existence judgments. Finally, to address this failure mode, we design a preference-data construction pipeline which leverages HIIs to build high-quality datasets without relying on any proprietary model (e.g., GPT-4~\cite{achiam2023gpt}).

\begin{figure}[htbp]
    \centering
    \resizebox{0.95\columnwidth}{0.95\height}{\includegraphics[trim=15em 8em 12em 5em,clip,width=1\columnwidth]{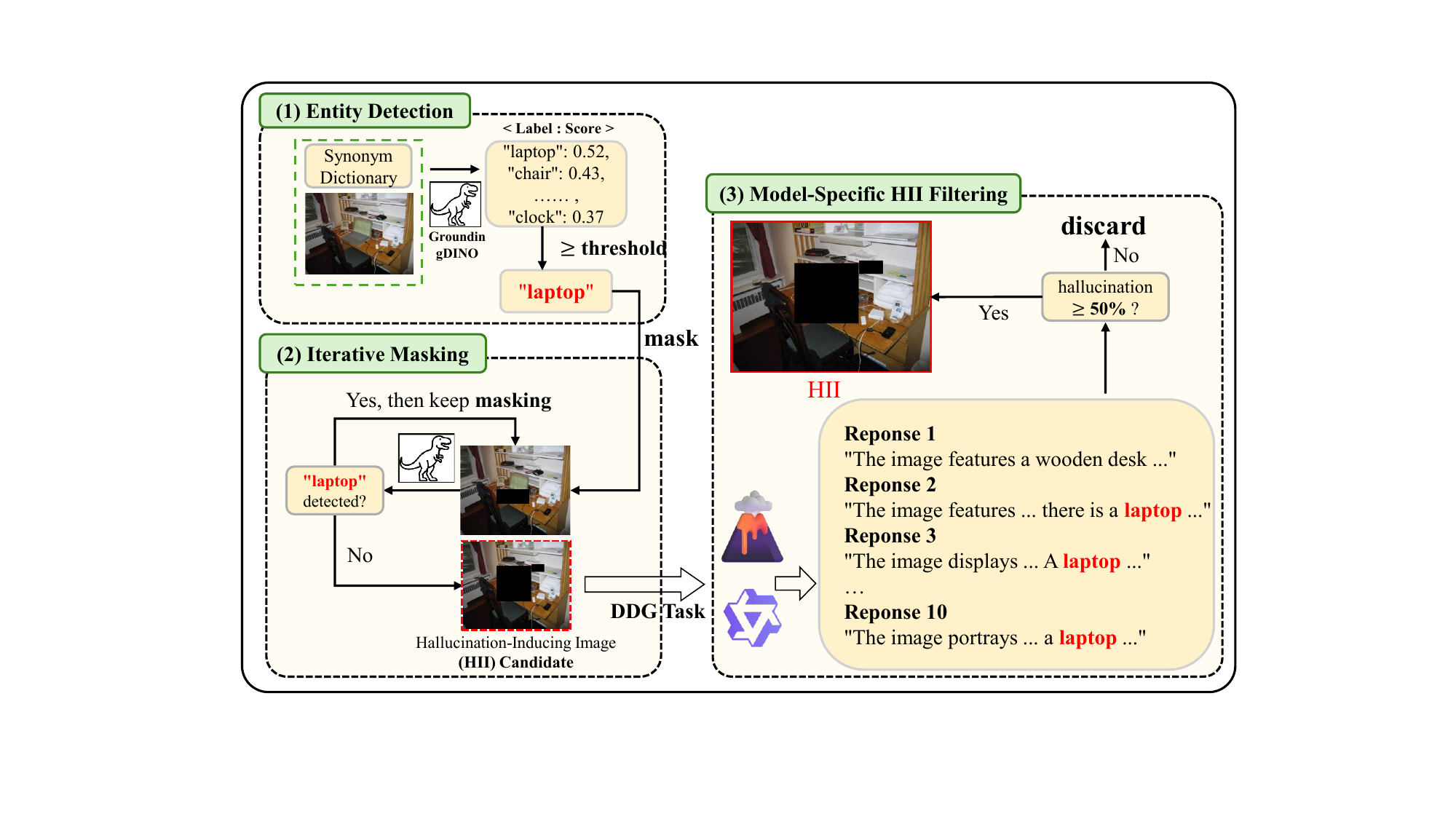}} 
    \caption{\textbf{Overview of the HII Synthesis Pipeline.} \textbf{(1)}: Potential entities are identified via GroundingDINO based on a predefined synonym dictionary. \textbf{(2)}: A detection-masking cycle is employed to achieve complete occlusion of the target entity, yielding HII candidates. \textbf{(3)} Task the target VLM to perform DDG and retain only those images with HR $\ge$ 50\%.}
    \label{fig::HIIs_pipeline}
\end{figure}

\subsection{Hallucination-Inducing Images (HIIs) Synthesis}
In this section, we describe the pipeline~\autoref{fig::HIIs_pipeline} designed to synthesize and select Hallucination-Inducing Images (HIIs). To support open-vocabulary detection, we utilize GroundingDINO ~\cite{liu2024grounding} to verify the presence of entities within an image. However, since the space of potential object categories is prohibitively large, we constrain our research scope to the 80 MS-COCO categories to ensure reproducibility and detection precision. 

\vspace*{-11px}

\paragraph{Entity Detection.}\label{sub::initial_detection} Following the methodology of ~\cite{rohrbach2018object}, we utilize a synonym dictionary to map detected entities--specifically those recognized within the dictionary--onto 80 target classes. Candidate objects are then identified for each class. To ensure robust detection, we only retain target classes whose confidence scores exceed a predefined threshold (e.g., 0.5). More examples of the synonym dictionary are presented in~\autoref{sub::synonym_dict}.

\vspace*{-11px}
\paragraph{Iterative Masking.} Next, a targeted masking strategy is applied to each filtered class. To mitigate issues related to incomplete occlusion or detection gaps, we implement an iterative masking procedure: for a given (image, class) pair, we perform successive detection-and-masking cycles until the target entity is fully occluded. Consequently, a single source image may yield multiple masked variants. 

\vspace*{-11px}
\paragraph{Model-Specific HII Filtering.} Finally, the synthesized masked images are utilized to task the target VLMs to perform Detailed Description Generation (DDG) tasks. To ensure the robustness of our findings, we employ sampling-based decoding to generate 10 independent responses for each masked image. We then extract the object entities mentioned in these responses using language toolkits~(e.g., nltk). An image is categorized as an HII specific to a model if that model asserts the presence of the successfully masked object in a majority of its responses (i.e., at least 5 out of 10). By requiring this frequency threshold, we filter out stochastic noise and ensure that the identified hallucinations are consistent model-specific behaviors directly tied to the absence of visual evidence.

\begin{figure}[htbp]
    \centering
    \resizebox{0.90\columnwidth}{0.90\height}{\includegraphics[trim=20em 15em 22em 0em,clip,width=0.9\columnwidth]{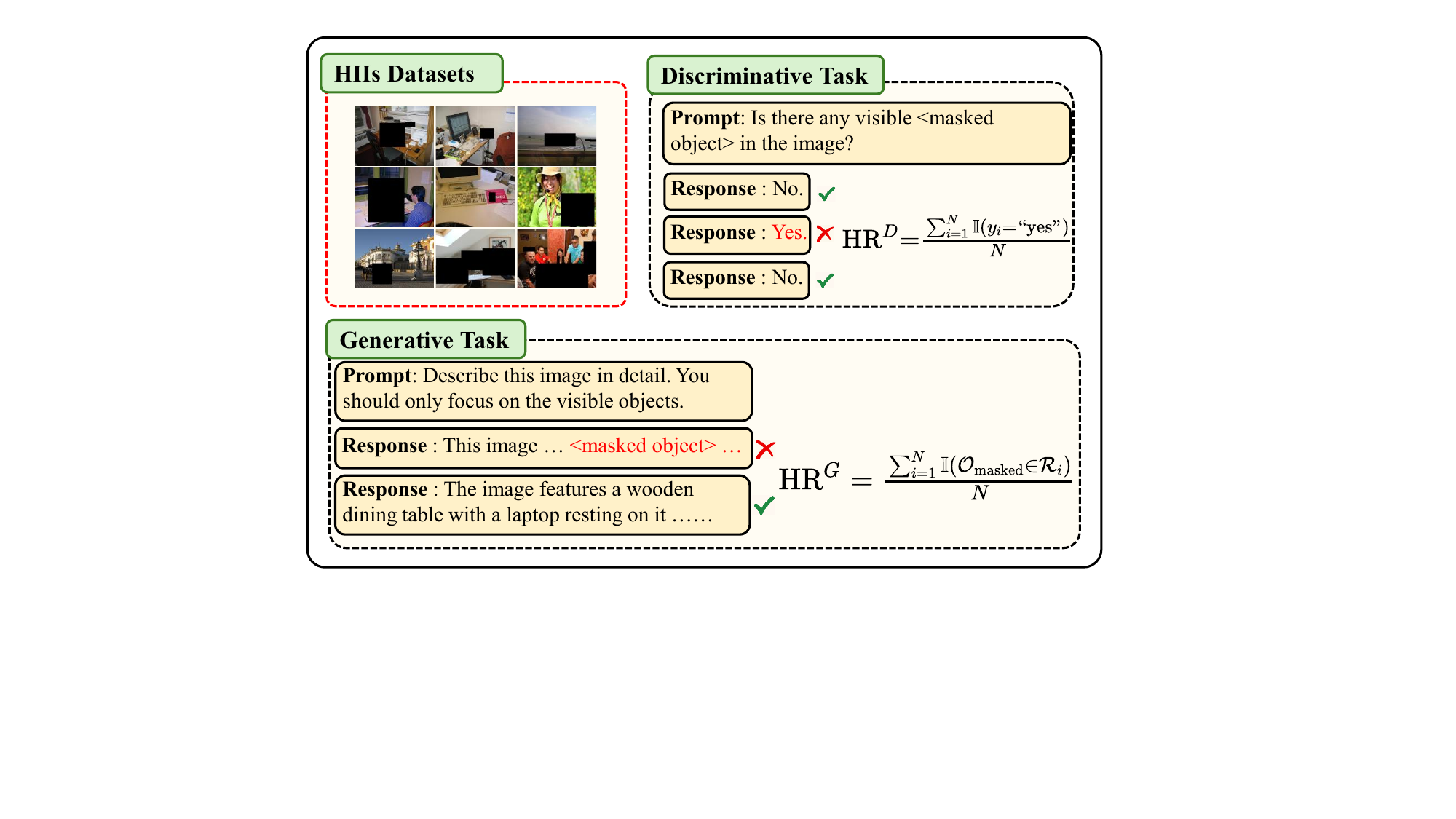} }
    \caption{\textbf{Masked-Object-Hallucination Benchmark.}}
    \label{fig:MOH_Bench}
\end{figure}

\subsection{The Masked-Object-Hallucination Benchmark}
\label{sub::MOH_benchmark}
Using the HIIs synthesis pipeline, we curate the \textbf{Masked-Object Hallucination Benchmark (MOH)} in~\autoref{fig:MOH_Bench}. We collect model-specific HIIs from a diverse set of VLMs, including the Qwen-VL family~\cite{wang2024qwen2vlenhancingvisionlanguagemodels,bai2025qwen25vltechnicalreport,bai2023qwenvlversatilevisionlanguagemodel} and LLaVA-1.5~\cite{liu2023visualinstructiontuning} To identify the most challenging samples, we take the \textbf{intersection} of these model-specific HII datasets, selecting images that consistently trigger hallucinations across all tested architectures to form our core evaluation set.

\vspace*{-10px}
\paragraph{Evaluation Tasks.}
\label{sub::MOH_eval} To evaluate VLM performance comprehensively, we design two distinct evaluation paradigms:
\begin{itemize}
    \vspace*{-11px}
    \item \textbf{Discriminative Probing:} We present the model with a direct query: \textit{“Is there any visible [masked object] in the image?”}. A “No” response is categorized as correct, while a “Yes” indicates a hallucination.
    \vspace*{-5px}
    \item \textbf{Generative Description:} We task the model with DDG and extract entities from the generated text. Specifically, we identify all mentioned entities and normalize them using our synonym dictionary. A hallucination is recorded if the model explicitly claims the presence of the masked object.
\end{itemize}

\vspace*{-12px}
\paragraph{Scene Taxonomy.} Since our MOH benchmark comprises the intersection of HIIs identified by models with diverse scales and architectures, it serves as a robust instrument for investigating universal hallucination patterns. To further uncover these underlying contextual drivers, we manually partition the benchmark into ten distinct environmental domains: \textit{Waterfront, Street, Railroad, Office, Dining Room, Kitchen, Bathroom, Ski Resort, Other Outdoor,} and \textit{Other Indoor}. This taxonomy facilitates a granular analysis of the scene-conditioned hallucination pattern. We provide quantitative statistics on scene-object co-occurrence in~\autoref{sec::results_MOH_benchmark}.

\vspace*{-11px}
\paragraph{Metric} Unlike the standard CHAIR$_i$ metric ~\cite{rohrbach2018object}, which can be sensitive to the frequency of correctly identified objects and prone to inflation via redundant mentions, we utilize the \textbf{Hallucination Rate (HR)} as our primary evaluation metric. For the generative task, we define HR as the ratio of model responses containing the masked object to the total number of generated responses:

\vspace*{-15px}
\begin{equation}
    \text{HR}^G = \frac{\sum_{i=1}^{N} \mathbb{I}(\mathcal{O}_{\text{masked}} \in \mathcal{R}_i)}{N}
\end{equation}

where $N$ denotes the total number of evaluation responses, $\mathcal{R}_i$ represents the set of objects extracted from the $i$-th response $y_i$, and $\mathbb{I}(\cdot)$ is the indicator function that outputs 1 if the masked object $\mathcal{O}_{\text{masked}}$ is detected in the response. Similarly, we can define HR for the discriminative task as follows:

\vspace*{-10px}
\begin{equation} \text{HR}^D = \frac{\sum_{i=1}^{N} \mathbb{I}(y_i = \text{“yes”})}{N} 
\end{equation} 

Since our benchmark specifically focuses on the “negative existence” of a known masked object, HR provides a more direct and precise measure of a VLM's susceptibility to hallucinatory claims without the noise introduced by the density of other present objects.

\begin{figure}[htbp]
    \centering
    \includegraphics[trim=17em 13em 12em 15em,clip,width=1\columnwidth]{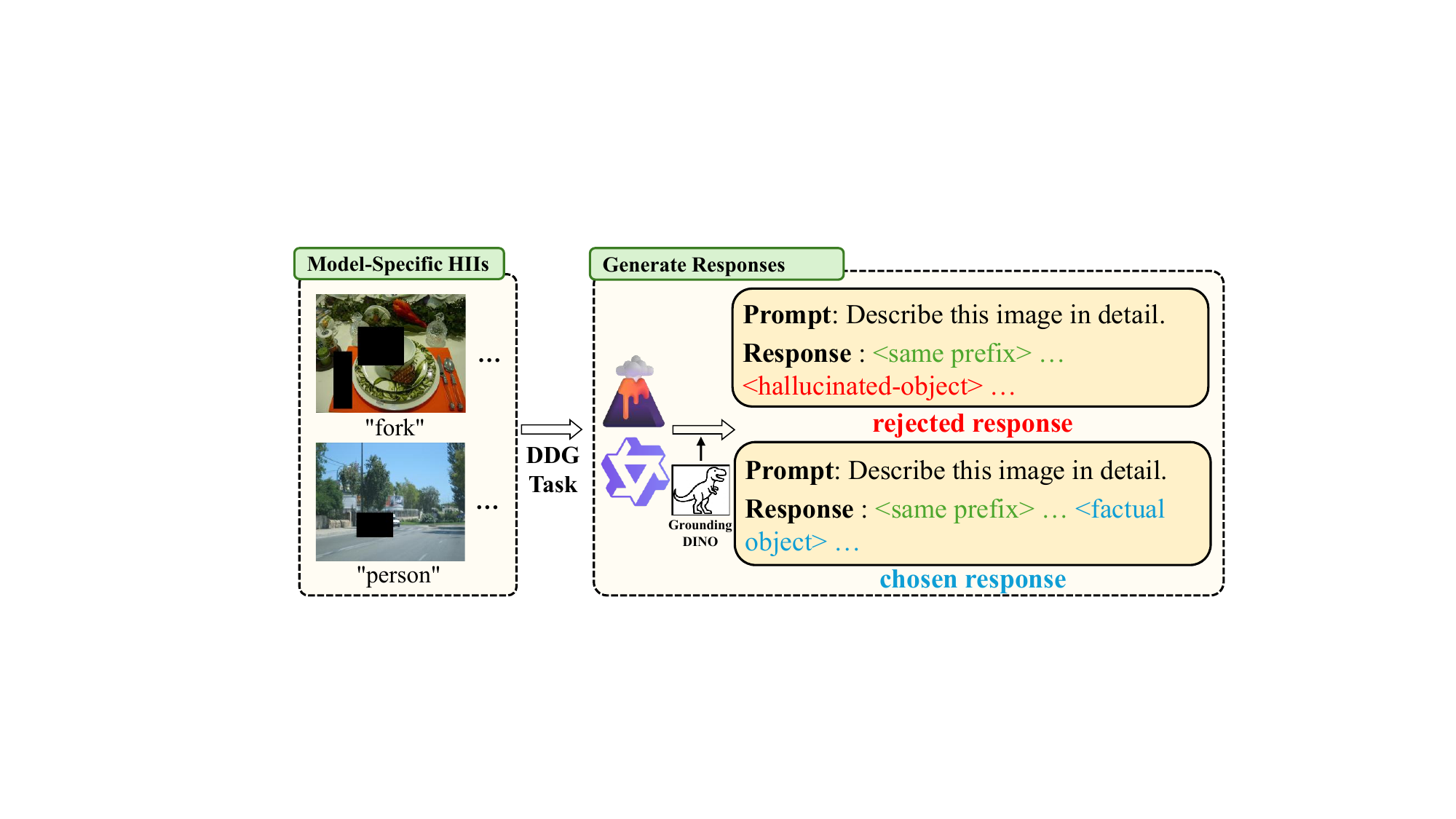} 
    \caption{\textbf{Preference Dataset Generation.} Utilizing curated model-specific HIIs, we construct contrastive response pairs where the \textit{chosen} and \textit{rejected} responses share an identical prefix. Verified by GroundingDINO, the rejected responses contain the hallucinated objects, whereas the chosen response describes only factual entities.}
    \label{fig::preference_pipeline}
\end{figure}

\subsection{Preference Dataset Construction for DPO}
To further mitigate hallucinatory behaviors, we leverage the curated model-specific HII datasets to construct a preference dataset for {Direct Preference Optimization (DPO)} as shown in~\autoref{fig::preference_pipeline}. To ensure the preference pairs remain within the model's internal distribution, we directly utilize the target VLMs to generate both chosen and rejected responses. A synonym dictionary is employed to detect potential entity words within the generated responses as introduced in~\autoref{sub::MOH_eval}. Subsequently, \textbf{GroundingDINO} ~\cite{liu2024grounding} is utilized to verify the physical presence of these entities in the corresponding image. If GroundingDINO fails to detect the extracted entity, the response is confirmed to contain a hallucination and is categorized as “rejected”.

However, standard DPO on full-length responses often introduces noise, as a “rejected” response may still contain significant factual information. Drawing inspiration from {SENTINEL} ~\cite{peng2025mitigatingobjecthallucinationssentencelevel}, we adopt a sentence-by-sentence generation strategy to enforce fine-grained alignment. Specifically, we identify pairs where the chosen and rejected responses share an identical prefix and diverge only at a specific sentence. In these pairs, the “chosen sentence” is factually accurate, while the “rejected sentence” contains at least one hallucinated physical entity. By optimizing on these minimal contrastive pairs, we ensure that the DPO training focuses strictly on the hallucinated sentences, effectively preventing the VLM from learning from noisy supervision signals while preserving its underlying factual knowledge. 
By integrating Direct Preference Optimization (DPO) with the concept of Hallucination-Inducing Images (HII), we propose \textbf{HII-DPO}, which is formulated as follows:

{\small
\begin{align}
\mathcal{L}(\boldsymbol{\theta}) = &-\mathbb{E}_{(\boldsymbol{x}, \boldsymbol{v}' \boldsymbol{y}^+, \boldsymbol{y}^-) \sim \mathcal{D}} \bigg[ \log \sigma \bigg( \beta \log \frac{\pi_{\boldsymbol{\theta}}(\boldsymbol{y}^+ | \boldsymbol{x,v'})}{\pi_{\text{ref}}(\boldsymbol{y}^+ | \boldsymbol{x,v'})} \nonumber \\
&\qquad \qquad \qquad \qquad - \beta \log \frac{\pi_{\boldsymbol{\theta}}(\boldsymbol{y}^- | \boldsymbol{x,v'})}{\pi_{\text{ref}}(\boldsymbol{y}^- | \boldsymbol{x,v'})} \bigg) \bigg]
\end{align}
}
Where $\boldsymbol{y}^+$ and $\boldsymbol{y}^-$ represent the generated chosen response and rejected response, respectively. $\boldsymbol{x}$ denotes the context or prompt, and $\boldsymbol{v}'$ refers to the Hallucination-Inducing Image (HII). The HII is defined as $\boldsymbol{v}' = \boldsymbol{v} - \boldsymbol{o}$, where $\boldsymbol{v}$ is the original natural image and $\boldsymbol{o}$ represents the specific object that has been removed (or masked out). We present a mathematical explanation about why HIIs are effective for preference alignment in mitigating object hallucination in~\autoref{sub::math_explanation}.

\vspace*{-11px}

\vspace{5pt}
\section{Experiments}

\label{sec::experiments}
In this section, we conduct extensive experiments to demonstrate the superiority of HII-DPO. Our method achieves state-of-the-art performance across multiple hallucination benchmarks while maintaining competitive general capabilities. We first detail the experimental setup in \autoref{sec::experimental_setups}, followed by the presentation of main results in \autoref{sec::main_results}. In~\autoref{sec::results_MOH_benchmark}, we evaluate existing approaches on our proposed MOH benchmark to verify the scene-conditioned hallucination pattern and analyze the correlation between scenes and objects. Furthermore, we provide an ablation study in \autoref{sec::ablation study}. Finally, inspired by mDPO \cite{wang2024mdpoconditionalpreferenceoptimization}, we incorporate visual contrastive signals for preference alignment and compare our approach with mDPO and Re-Align \cite{xing2025realignaligningvisionlanguage}, highlighting potential limitations of this training paradigm.

\subsection{Experimental Setups}
\label{sec::experimental_setups}
\paragraph{Dataset} 
Following the HII Synthesis Pipeline in~(\autoref{fig::HIIs_pipeline}), we curated a final set of approximately 800 HIIs to constitute the MOH benchmark. For the preference dataset, we identified model-specific HIIs tailored to various architectures and scales, subsequently constructing chosen-rejected preference pairs using these HIIs. To prevent data leakage, we utilized distinct sets of base images for the synthesis of HIIs which are for the MOH benchmark and the preference dataset construction, respectively. Specifically, for MOH, we chose COCO2014~\cite{lin2015microsoftcococommonobjects} as the base dataset. The preference dataset was synthesized based on 8,000 natural images sampled from Visual Genome~\cite{krishna2017visual}. To verify the physical existence of entities mentioned in the generated responses, we employed GroundingDINO~\cite{liu2024grounding} as our open-vocabulary detection model.

\vspace*{-10px}
\paragraph{Evaluation Benchmarks.} For hallucination benchmarks, we use Object Hallucination~\cite{rohrbach2018object}, Amber~\cite{wang2024amberllmfreemultidimensionalbenchmark} and HallusionBench~\cite{guan2024hallusionbench}. For general VQA benchmarks, we utilize ScienceQA~\cite{lu2022learn}, TextVQA~\cite{singh2019towards}, VQAv2~\cite{goyal2017making} and MM-Vet~\cite{yu2023mm}. 

\vspace*{-10px}
\paragraph{Baseline Approaches.}
Following the standard configurations established in recent studies, we adopt LLaVA-v1.5-7B and 13B \cite{liu2023visualinstructiontuning} as the base architectures for our experiments. We conduct comprehensive comparisons between our approach and several state-of-the-art VLM alignment frameworks including Re-Align~\cite{xing2025realignaligningvisionlanguage}, SENTINEL~\cite{peng2025mitigatingobjecthallucinationssentencelevel}, HA-DPO~\cite{zhao2024hallucinationsenhancinglvlmshallucinationaware}, RLAIF-V~\cite{yu2025rlaif}, TPO~\cite{he2024topic}, POVID~\cite{zhou2024aligningmodalitiesvisionlarge}, HSA-DPO~(including Vanilla DPO)~\cite{xiao2025detectingmitigatinghallucinationlarge} and mDPO~\cite{wang2024mdpoconditionalpreferenceoptimization}. All baseline results are reproduced using publicly available model weights or the provided preference datasets.

\begin{table*}[t]
\centering
\caption{\textbf{Comparison with state-of-the-art methods on hallucination and general benchmarks.} Best results are \textbf{bold} and second best are \underline{underlined}. 
}
\label{tab:main_results} 
\resizebox{\textwidth}{!}{
\begin{tabular}{llcccccc|cccc|c}
\toprule
\multirow{3}{*}{\textbf{Model}} & \multirow{3}{*}{\textbf{Method}} & \multicolumn{6}{c|}{\textbf{Hallucination benchmarks}} & \multicolumn{4}{c|}{\textbf{General benchmarks}} & \multirow{3}{*}{\textbf{Avg Rank}} \\ 
\cmidrule(lr){3-8} \cmidrule(lr){9-12}
 & & \multicolumn{2}{c}{\textbf{Object HalBench}} & \multicolumn{3}{c}{\textbf{AMBER}} & \textbf{HallusionBench} & \textbf{VQAv2} & \textbf{TextVQA} & \textbf{ScienceQA} & \textbf{MM-Vet} \\
\cmidrule(lr){3-4} \cmidrule(lr){5-7} \cmidrule(lr){8-8} \cmidrule(lr){9-9} \cmidrule(lr){10-10} \cmidrule(lr){11-11} \cmidrule(lr){12-12}
 & & CHAIR\_s. $\downarrow$ & CHAIR\_i. $\downarrow$ & CHAIR $\downarrow$ & Hal. $\downarrow$ & Cog. $\downarrow$ & qAcc. $\uparrow$ & Acc. $\uparrow$ & Acc. $\uparrow$ & Image Acc. $\uparrow$ & Overall $\uparrow$ \\
\midrule
\multirow{9}{*}{LLaVA-v1.5-7B} 
 & baseline & 52.7 & 28.0 & 8.4 & 35.5 & 4.0 & 46.86 & \textbf{78.5} & \textbf{58.2} & 66.8 & 31.0 & 6.50 \\
 & Re-Align~\cite{xing2025realignaligningvisionlanguage} & 36.2 & 16.3 & 5.0 & 26.8 & 1.9 & \underline{47.62} & 71.9 & 48.0 & 66.6 & 28.8 & 6.35 \\
 & mDPO~\cite{wang2024mdpoconditionalpreferenceoptimization} & 30.7 & 16.0 & 5.0 & 27.5 & 2.4 & 46.15 & - & 47.3 & 67.3 & - & 6.18 \\
 & HA-DPO~\cite{zhao2024hallucinationsenhancinglvlmshallucinationaware} & 37.0 & 20.9 & 6.7 & 30.9 & 3.3 & \textbf{47.74} & 77.6 & \underline{56.7} & \textbf{69.7} & 30.6 & 5.55 \\
 & POVID~\cite{zhou2024aligningmodalitiesvisionlarge} & 33.4 & 16.6 & 5.3 & 28.7 & 3.0 & 46.59 & 77.2 & 56.6 & 68.8 & 31.8 & 5.60 \\
 & RLAIF-V~\cite{yu2025rlaif} & 7.8 & 4.2 & \underline{2.8} & 15.7 & \textbf{0.9} & 35.43 & 75.2 & 55.1 & 68.2 & 29.9 & 4.85 \\
 & TPO~\cite{he2024topic} & 5.6 & 3.2 & 3.6 & 20.5 & 1.6 & 40.12 & 75.9 & 55.3 & 67.1 & 25.7 & 5.30 \\
 & SENTINEL~\cite{peng2025mitigatingobjecthallucinationssentencelevel} & \underline{5.5} & \underline{3.0} & 2.9 & \underline{14.6} & \underline{1.2} & 47.56 & \underline{78.4} & \textbf{58.2} & \underline{69.2} & \underline{32.6} & \underline{2.40} \\
 & \textbf{HII-DPO} & \textbf{4.0} & \textbf{2.5} & \textbf{2.7} & \textbf{13.5} & \textbf{0.9} & \textbf{47.74} & 77.8 & \textbf{58.2} & 68.6 & \textbf{33.8} & \textbf{1.70} \\
 \midrule
\multirow{7}{*}{LLaVA-v1.5-13B} 
 & baseline & 46.0 & 23.0 & 6.9 & 31.9 & 3.3 & 46.43 & \textbf{80.0} & \textbf{61.2} & 71.6 & \underline{36.0} & 4.75 \\
 & Re-Align~\cite{xing2025realignaligningvisionlanguage} & 38.4 & 19.3 & 6.1 & 31.8 & 2.4 & 46.43 & 76.3 & 53.3 & 69.0 & 33.5 & 5.94 \\
 & mDPO~\cite{wang2024mdpoconditionalpreferenceoptimization} & 33.3 & 16.6 & 4.6 & 25.2 & 2.0 & 46.23 & - & 55.7 & 69.4 & - & 5.31 \\
 & vanilla-DPO~\cite{li2024hsadpo} & 6.7 & 3.6 & 2.8 & 15.5 & 1.6 & 46.41 & 79.2 & 60.4 & \underline{71.8} & 35.0 & 4.05 \\
 & HSA-DPO~\cite{li2024hsadpo} & 5.0 & 3.1 & \textbf{2.0} & 12.1 & 1.1 & 46.14 & 78.0 & 60.0 & 71.3 & 34.0 & 4.00 \\
 & SENTINEL~\cite{peng2025mitigatingobjecthallucinationssentencelevel} & \underline{4.5} & \underline{2.4} & 2.7 & \underline{11.9} & \underline{0.9} & \underline{46.77} & \underline{79.9} & \underline{61.0} & \textbf{72.8} & \textbf{36.2} & \underline{2.00} \\
 & \textbf{HII-DPO} & \textbf{2.9} & \textbf{1.7} & \underline{2.3} & \textbf{11.4} & \textbf{0.6} & \textbf{46.86} & 79.5 & 60.8 & \textbf{72.8} & \textbf{36.2} & \textbf{1.60} \\
\bottomrule
\end{tabular}
}
\end{table*}


\subsection{Main Results}
\label{sec::main_results}
In \autoref{tab:main_results}, we present a comprehensive comparison of various vision-language alignment frameworks across both hallucination-oriented and general VQA benchmarks, using LLaVA-v1.5-7B and LLaVA-v1.5-13B~\cite{liu2023visualinstructiontuning} as backbone models. Experimental results demonstrate that our method achieves state-of-the-art performance across multiple evaluation datasets. Specifically, for the LLaVA-v1.5-7B model, our method significantly reduces hallucination rates compared to the baseline, achieving a maximum reduction of up to $92\%$. Compared to the previous leading method SENTINEL~\cite{peng2025mitigatingobjecthallucinationssentencelevel}, our approach yields average performance gains of $27.3\%$ on Obj-Hal and $13.1\%$ on AMBER. Furthermore, our method consistently achieves SOTA results on hallucination benchmarks for the 13B model, underscoring its robust generalizability on larger model scales. Regarding general VQA benchmarks, while a marginal performance trade-off is observed in certain metrics compared to the baseline, such a phenomenon is expected in hallucination mitigation approaches. As discussed in our mathematical explanation, the suppression of over-reliance on language priors can inherently interfere with the VLM's internal knowledge base. Nevertheless, our approach still maintains highly competitive performance on general VQA tasks, even achieving the best results in several categories.

\begin{figure*}[htbp] 
    \centering
    \includegraphics[width=0.95\textwidth]{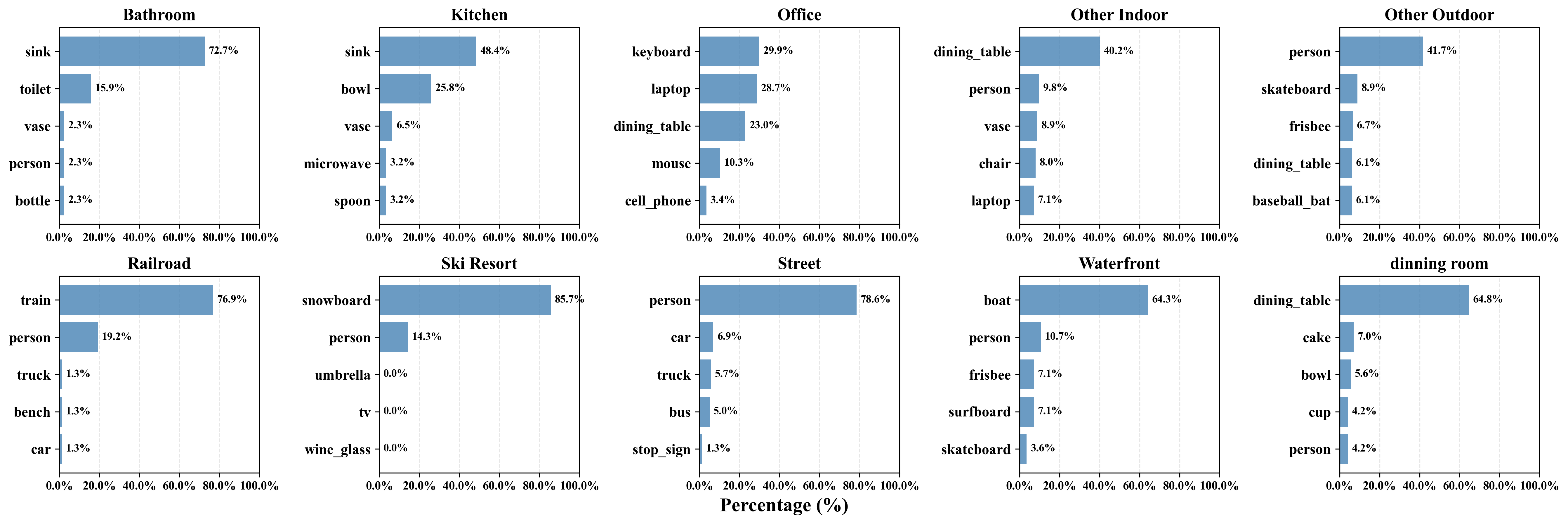} 
    \caption{\textbf{Distribution of the top-5 masked objects across ten environmental settings.} Each subplot illustrates the percentage of specific objects that trigger hallucinations within the scenario. These statistics reveal the ingrained scene-conditioned hallucination pattern within VLMs; for instance, “boat” dominates in waterfront context, while “train” is the primary driver of hallucinations in the railroad setting.}
    \label{fig::co_occurence}
\end{figure*}

\begin{table}[htbp]
    \centering
    \caption{\textbf{Results on the MOH benchmark for LLaVA-v1.5-7B.} We report the hallucination rates HR$^D$ and HR$^G$}
    \label{tab::MOH_7B}
    \resizebox{\columnwidth}{!}{
    \begin{tabular}{l|cccccccc}
        \toprule
        \textbf{Metric} & \textbf{Baseline} & \textbf{Re-Align} & \textbf{mDPO} & \textbf{POVID} & \textbf{HA-DPO} & \textbf{TPO} & \textbf{SENTINEL} & \textbf{Ours} \\ 
        \midrule
        HR$^D$ $\downarrow$ & 52.2 & 49.9 & 50.7 & 51.1 & 45.2 & 31.7 & 38.3 & \textbf{28.1} \\ 
        HR$^G$ $\downarrow$ & 68.6 & 49.0 & 53.2 & 65.1 & 63.4 & 36.6 & 38.8 & \textbf{18.8} \\ 
        \bottomrule
    \end{tabular}
    }
\end{table}

\begin{table}[htbp]
    \centering
    \caption{\textbf{Results on the MOH benchmark for LLaVA-v1.5-13B.}}
    \label{tab::MOH_13B}
    \resizebox{\columnwidth}{!}{
    \begin{tabular}{l|ccccccc}
        \toprule
        \textbf{Metric} & \textbf{Baseline} & \textbf{Re-Align} & \textbf{mDPO} &  \textbf{vanilla-DPO} & \textbf{HSA-DPO} & \textbf{SENTINEL} & \textbf{Ours} \\
        \midrule

        HR$^D$ $\downarrow$ & 69.5 & \textbf{39.4} & 45.6 & 46.2 & 45.3 & 56.8 & 43.8 \\ 

        HR$^G$ $\downarrow$ & 67.9 & 51.1 & 48.3 & 35.0 & 28.2 & 36.5 & \textbf{17.5} \\ 
        \bottomrule
    \end{tabular}
    }
\end{table}

\subsection{Results on MOH benchmark}
\label{sec::results_MOH_benchmark}

\paragraph{Statistics in HIIs.}
As introduced in~\autoref{sub::MOH_benchmark}, we curate the MOH benchmark by taking the intersection of multiple model-specific HII datasets, resulting in a collection of samples that are most prone to trigger VLM hallucinations. Consequently, by conducting a statistical analysis of the MOH dataset, we can identify common hallucination patterns across VLMs with various model scales and architectures. Furthermore, we can infer the specific contexts in which VLMs are likely to falsely claim the existence of particular objects. In~\autoref{fig::co_occurence}, we present the top-5 objects which are most frequently hallucinated by VLMs within specific scenes, yielding several noteworthy observations. For instance, VLMs exhibit a strong tendency to hallucinate a “sink” in bathroom settings, a “boat” in waterfront scenarios, and a “train” on railroads. These findings suggest that the scene-conditioned hallucination is rooted in VLMs. For more HII examples, please refer to~\autoref{sec::case_MOH}.

\paragraph{Evaluation on existing methods.} We evaluate prior state-of-the-art methods on our MOH benchmark and present the experimental results in \autoref{tab::MOH_7B} and \autoref{tab::MOH_13B}. For baseline models, {the generative task} (DDG) exhibits a higher Hallucination Rate (HR) compared to {the discriminative task}. We hypothesize that VLMs are more prone to hallucinating when generating verbose responses due to the {autoregressive nature} of the models, where errors tend to accumulate over successive time steps. Among existing approaches, average hallucination rates reach as high as 47.8\% and 49.4\% on discriminative and generative tasks, respectively, underscoring the persistence of scene-conditioned hallucinations in current VLMs.
In contrast, our method consistently reduces hallucination rates in both {discriminative} and {generative tasks}. Specifically, our approach achieves an average reduction in hallucination rates of 41.6\% for {the discriminative task} and 73.4\% for {the generative task}, demonstrating its effectiveness in suppressing these scene-conditioned hallucination patterns. Notably, we find that VCA-based methods \cite{xing2025realignaligningvisionlanguage,wang2024mdpoconditionalpreferenceoptimization} are particularly effective at mitigating hallucinations in \textbf{the discriminative task}; we provide a more in-depth discussion on this observation in~\autoref{sec::comparision_with_vision_contrastive_methods}.


\begin{table*}[htbp]
\centering
\caption{\textbf{Comparison with existing VCA-based methods on hallucination and general benchmarks.} We evaluate the integration of our HIIs with the SENTINEL framework. While the results show that {SENTINEL + HIIs} achieves a new state-of-the-art in hallucination suppression, it incurs an {inevitable performance degradation} in general understanding capabilities.}
\label{tab::contrastive_vision} 
\resizebox{\textwidth}{!}{
\begin{tabular}{llccccccc|cccc}
\toprule
\multirow{3}{*}{\textbf{Model}} & \multirow{3}{*}{\textbf{Method}} & \multicolumn{7}{c|}{\textbf{Hallucination benchmarks}} & \multicolumn{4}{c}{\textbf{General benchmarks}} \\ 
\cmidrule(lr){3-9} \cmidrule(lr){10-13}
 & & \multicolumn{2}{c}{\textbf{Object HalBench}} & \multicolumn{3}{c}{\textbf{AMBER}} & \multicolumn{2}{c|}{\textbf{MOH}} & \textbf{VQAv2} & \textbf{TextVQA} & \textbf{ScienceQA} & \textbf{MM-Vet} \\
\cmidrule(lr){3-4} \cmidrule(lr){5-7} \cmidrule(lr){8-9} \cmidrule(lr){10-10} \cmidrule(lr){11-11} \cmidrule(lr){12-12} \cmidrule(lr){13-13}
 & & CHAIR\_s. $\downarrow$ & CHAIR\_i. $\downarrow$ & CHAIR $\downarrow$ & Hal. $\downarrow$ & Cog. $\downarrow$ & HR$^D$. $\downarrow$ & HR$^G$. $\downarrow$ & Acc. $\uparrow$ & Acc. $\uparrow$ & Image Acc. $\uparrow$ & Overall $\uparrow$ \\
\midrule
\multirow{5}{*}{LLaVA-v1.5-7B} 
 & baseline & 52.7 & 28.0 & 8.4 & 35.5 & 4.0 & 52.2 & 68.6 & \textbf{78.5} & \textbf{58.2} & 66.8 & 31.0 \\
 & Re-Align & 36.2 & 16.3 & 5.0 & 26.8 & 1.9 & 49.9 & 49.0 & 71.9 & 48.0 & 66.6 & 28.8 \\
 & mDPO & 30.7 & 16.0 & 5.0 & 27.5 & 2.4 & 50.7 & 53.2 & - & 47.3 & 67.3 & - \\
& SENTINEL & 5.5 & 3.0 & 2.9 & 14.6 & 1.2 & 38.3 & 38.8 & 78.4 & \textbf{58.2} & \textbf{69.2} & \textbf{32.6} \\
& \textbf{SENTINEL + HIIs} & \textbf{1.5} & \textbf{1.0} & \textbf{1.6} & \textbf{6.7} & \textbf{0.4} & \textbf{13.5} & \textbf{8.6} & 77.0 & 57.5 & 67.9 & 32.2 \\

\bottomrule
\end{tabular}
}
\end{table*}

\subsection{Ablation Study}
\label{sec::ablation study}
We conduct a series of ablation studies to further analyze the effectiveness of HII-DPO using LLaVA-v1.5-7B as the baseline. We present more ablation studies in~\autoref{sub::more_ablation}.

\begin{table}[htbp]
\centering
\caption{\textbf{Ablation study on the detection threshold in the entity detection phase.}}
\label{tab::mask_threshold}
\resizebox{\columnwidth}{!}{\begin{tabular}{lcccccc}
\toprule
\multirow{2}{*}{\textbf{Threshold}} & \multicolumn{2}{c}{\textbf{Object HalBench}} & \multicolumn{3}{c}{\textbf{AMBER}} & \textbf{ScienceQA} \\ 
\cmidrule(lr){2-3} \cmidrule(lr){4-6} \cmidrule(lr){7-7}
 & CHAIR\_s. $\downarrow$ & CHAIR\_i. $\downarrow$ & CHAIR $\downarrow$ & Hal $\downarrow$ & Cog $\downarrow$ & Image Acc. $\uparrow$ \\ 
\midrule
Baseline & 52.7 & 27.9 & 8.4 & 35.5 & 4.0 & 66.8 \\
0.25 & 7.2 & 3.0 & 3.2 & 14.6 & 1.2 & 67.9 \\
0.35 & \textbf{4.0} & \textbf{2.5} & \textbf{2.7} & \textbf{13.5} & \textbf{0.9} & \textbf{68.6} \\
0.45 & 9.3 & 3.2 & 3.0 & 15.4 & 1.2 & 67.5 \\ 
\bottomrule
\end{tabular}}
\end{table}

\paragraph{Mask threshold.} In this part, we evaluate the impact of the confidence threshold within the HII Synthesis Pipeline in~\autoref{fig::HIIs_pipeline} and summarize the empirical results in \autoref{tab::mask_threshold}. We observe that a lower threshold often leads to the inaccurate detection of entity labels, introducing semantic ambiguity and noise into the preference pairs, which can ultimately exacerbate hallucinations. Conversely, an excessively high threshold may filter out valid labels and restrict the yield of HIIs, thereby diminishing the diversity of the training data. Overall, 0.35 is the optimal choice for balancing detection robustness with the preservation of sample diversity.

\vspace*{-18px}
\begin{table}[htbp]
\centering
\caption{\textbf{Ablation study on the iterative masking strategy.}}
\label{fig::interative}
\resizebox{\columnwidth}{!}{\begin{tabular}{lcccccc}
\toprule
\multirow{2}{*}{\textbf{Method}} & \multicolumn{2}{c}{\textbf{Object HalBench}} & \multicolumn{3}{c}{\textbf{AMBER}} & \textbf{ScienceQA} \\ 
\cmidrule(lr){2-3} \cmidrule(lr){4-6} \cmidrule(lr){7-7}
 & CHAIR\_s. $\downarrow$ & CHAIR\_i. $\downarrow$ & CHAIR $\downarrow$ & Hal $\downarrow$ & Cog $\downarrow$ & Image Acc. $\uparrow$ \\ 
\midrule
LLaVA-v1.5-7B & 52.7 & 27.9 & 8.4 & 35.5 & 4.0 & 66.8 \\
w.o. iterative & 9.2 & 3.6 & 3.6 & 15.6 & 1.2 & 67.2 \\
w. iterative & \textbf{4.0} & \textbf{2.5} & \textbf{2.7} & \textbf{13.5} & \textbf{0.9} & \textbf{68.6} \\
\bottomrule
\end{tabular}}
\end{table}

\paragraph{Iterative masking strategy.} We further evaluate the effectiveness of the iterative masking strategy within the HII Synthesis Pipeline (\autoref{fig::HIIs_pipeline}). As illustrated by the experimental results in \autoref{fig::interative}, the iterative approach consistently outperforms the single-pass masking baseline. Relying solely on an initial mask often leads to incomplete occlusion of the target entity. Consequently, the open-vocabulary detector may still identify remnants of the partially masked object during dataset generation, thereby introducing semantic ambiguity into the resulting preference pairs.

\subsection{Comparison with VCA-based Methods}
\label{sec::comparision_with_vision_contrastive_methods}
As previously discussed, mDPO\cite{wang2024mdpoconditionalpreferenceoptimization} pioneered the integration of Vision-Contrastive Alignment (VCA) datasets into the DPO framework. This approach can be formally defined as:

\begin{align}
\mathcal{L}(\boldsymbol{\theta}) = &-\mathbb{E}_{(\boldsymbol{x}, \boldsymbol{v}^+,\boldsymbol{v}^- \boldsymbol{y}^+) \sim \mathcal{D}} \bigg[ \log \sigma \bigg( \beta \log \frac{\pi_{\boldsymbol{\theta}}(\boldsymbol{y}^+ | \boldsymbol{x,v^+})}{\pi_{\text{ref}}(\boldsymbol{y}^+ | \boldsymbol{x,v^+})} \nonumber \\
&\qquad \qquad \qquad \qquad - \beta \log \frac{\pi_{\boldsymbol{\theta}}(\boldsymbol{y}^+ | \boldsymbol{x,v^-})}{\pi_{\text{ref}}(\boldsymbol{y}^+ | \boldsymbol{x,v^-})} \bigg) \bigg]
\end{align}

Here, $\boldsymbol{v}^+$ denotes the original natural image, while $\boldsymbol{v}^-$ represents a visually corrupted or irrelevant image. $\boldsymbol{y}^+$ is the chosen response paired with $\boldsymbol{v}^+$. As shown in \autoref{tab::MOH_7B} and \autoref{tab::MOH_13B}, VCA-based methods effectively reduce hallucination rates on the MOH benchmark. Motivated by these findings, we leverage HIIs as the negative vision-contrastive signals ($\boldsymbol{v}^-$) and the corresponding unmasked natural images as the positive signals ($\boldsymbol{v}^+$) to investigate the efficacy of HIIs within the VCA paradigm. Specifically, we augmented the original training dataset provided by SENTINEL~\cite{peng2025mitigatingobjecthallucinationssentencelevel} with our HII samples and conducted extensive evaluations. As shown in \autoref{tab::contrastive_vision}, incorporating HIIs as vision-contrastive signals drastically mitigates hallucination rates across all evaluated benchmarks, including MOH. However, this reduction comes at the expense of general understanding capabilities---a trade-off we further analyze in the following part. As previously discussed in~\autoref{sec::intro}, DPO inherently decreases the probability of the chosen response $\boldsymbol{y}^+$ when paired with the negative image $\boldsymbol{v}^-$, even if portions of $\boldsymbol{y}^+$ contain semantic details that remain grounded in $\boldsymbol{v}^-$. This leads to two primary consequences: \textbf{(1) Conservative Model Behavior:} Since the DPO objective suppresses the likelihood of $\boldsymbol{y}^+$ when paired with the negative image $\boldsymbol{v}^-$, even the semantic elements that remain factually grounded in $\boldsymbol{v}^-$ are inevitably penalized. This often leads to an \textit{over-correction} effect, where the VLM adopts a more conservative generation strategy. Consequently, the model exhibits an increased tendency toward brevity, becoming more reluctant to provide detailed descriptions or venture into fine-grained inferences for ambiguous visual regions to minimize the risk of penalty. \textbf{(2) Erosion of the Internal Knowledge Base:} While the Vision-Contrastive Alignment (VCA) paradigm effectively disrupts harmful linguistic biases, it may simultaneously erode the model's internal knowledge base acquired during the Supervised Fine-Tuning (SFT) phase. This is primarily attributed to the {unconditional penalization} of the chosen response $\boldsymbol{y}^+$ when paired with the negative image $\boldsymbol{v}^-$. The resulting suppression of these established priors leads to a degradation in performance on general VQA tasks, where the model previously relied on such internal knowledge to bridge visual gaps.

\section{Conclusion}
Driven by the scene-object co-occurrence, we propose a systematic framework to investigate and mitigate the scene-conditioned hallucination pattern. We design a pipeline to synthesize model-specific Hallucination-Induced Images and subsequently extract their intersection to uncover universal hallucination patterns shared across various model architectures and scales. Based on HIIs, we introduce the MOH benchmark to rigorously evaluate a VLM's susceptibility to language bias. Furthermore, we leverage HIIs to construct preference datasets and employ HII-DPO to perform fine-grained preference alignment. Our experimental results demonstrate that this approach significantly mitigates hallucinations while preserving the general capabilities of the models.

\section*{Impact Statement}
Our work proposes HII-DPO a novel framework to mitigate scene-conditioned hallucinations in Vision-Language Models (VLMs). By prioritizing visual grounding over linguistic shortcuts, our method significantly enhances the reliability of AI. This is particularly critical in domains such as medical diagnostics and autonomous navigation, where factual accuracy and robust verification are indispensable for operational safety. Furthermore, our approach helps identify and suppress ingrained language biases, promoting more objective multimodal reasoning. Additionally, the proposed HII pipeline serves as a robust diagnostic instrument for the research community to systematically investigate and address model vulnerabilities.

\bibliography{main}
\bibliographystyle{icml2026}

\newpage
\appendix
\onecolumn

\section{Mathematical Explanation}
\label{sub::math_explanation}
We provide a probabilistic view of why \emph{HIIs} are effective for preference alignment in mitigating object hallucination.
For a given object $o$, define:
\begin{itemize}
    \item $v$: the original image (where $o$ is truly present).
    \item $v' = v - o$: a hallucination-inducing image where $o$ is removed (or masked out).
    \item $x$: the textual context providing an accurate but partial description of the input image.
    \item $z \in \{0,1\}$: a binary variable indicating whether the model \emph{claims} that $o$ is present in the image ($z=1$) or absent ($z=0$).
    \item $p_{\theta}(\cdot)$: the VLM parameterized by $\theta$.
\end{itemize}

We focus on the model's propensity to assert the presence of $o$ given $(x,v)$:
\begin{equation}
p_\theta(z=1 \mid x, v).
\label{eq:posterior_z1}
\end{equation}
By Bayes' rule,
\begin{align}
p_\theta(z=1 \mid x, v)
&= \frac{p_\theta(v \mid z=1, x)\, p_\theta(z=1 \mid x)}{p_\theta(v \mid x)} \nonumber\\
&\propto p_\theta(v \mid z=1, x)\, p_\theta(z=1 \mid x),
\label{eq:bayes_basic}
\end{align}
where $p_\theta(v\mid x)$ is the normalization term for \emph{fixed} $(x,v)$.
In this formulation, $p_\theta(z=1\mid x)$ captures the \emph{language prior}, representing the model's inherent expectation of object $o$ given the textual context $x$. And $p_\theta(v\mid z=1,x)$ is interpreted as the \emph{visual likelihood}, which quantifies the compatibility between the observed visual evidence and the hypothesis of $o$'s presence. Although standard VLMs do not explicitly parameterize these components, they are implicitly optimized during our preference alignment process to achieve a better balance between textual priors and visual evidence.

A more revealing form is obtained by considering \emph{posterior odds}. Using the identity
\[
\frac{p_\theta(z=1\mid x,v)}{p_\theta(z=0\mid x,v)}
=
\frac{p_\theta(v\mid z=1,x)}{p_\theta(v\mid z=0,x)}
\cdot
\frac{p_\theta(z=1\mid x)}{p_\theta(z=0\mid x)},
\]
we get the log-odds decomposition:
\begin{equation}
\begin{aligned}
\log \frac{p_\theta(z=1\mid x,v)}{p_\theta(z=0\mid x,v)}
&=
\underbrace{\log \frac{p_\theta(v\mid z=1,x)}{p_\theta(v\mid z=0,x)}}_{\text{visual evidence (likelihood ratio)}}
\\
&\quad +
\underbrace{\log \frac{p_\theta(z=1\mid x)}{p_\theta(z=0\mid x)}}_{\text{language prior (prior odds)}}.
\end{aligned}
\label{eq:odds_decomp}
\end{equation}

\paragraph{Why hallucinations happen after SFT.}
In many caption/instruction-tuning corpora, objects co-occur in stereotypical ways.
As a result, after SFT the prior odds can become highly skewed, i.e.,
$p_\theta(z=1\mid x) \gg p_\theta(z=0\mid x)$ for certain objects, even when the image does not support them.
Then the second term in \autoref{eq:odds_decomp} can dominate, pushing the model to claim $z=1$ (hallucination) despite weak or contradictory visual evidence.

\paragraph{Why HIIs help.}
Now consider the edited image $v'=v-o$, where $o$ is absent. The desired behavior is to increase
\begin{equation}
p_\theta(z=0\mid x,v')
\propto p_\theta(v' \mid z=0,x)\, p_\theta(z=0\mid x).
\label{eq:z0_vprime}
\end{equation}
Crucially, $v'$ often preserves the {contextual cues} that correlate with $o$ (scene, co-occurring objects, typical layouts), so the language prior may still favor $z=1$.
Thus, $v'$ creates {hard counterexamples} where the model must rely on visual grounding (negative evidence) instead of the scene-object co-occurrence heuristics.

In preference alignment, we provide pairs consisting of: (i) a \textit{chosen} response $y^{+}$ that correctly omits any mention of object $o$, and (ii) a \textit{rejected} response $y^{-}$ that falsely claims the existence of $o$ for the same $(x, v')$.
Optimizing the preference objective increases the model's probability of producing $y^{+}$ over $y^{-}$, which effectively increases the posterior odds:~$p_\theta(z=0\mid x,v')$.
Viewed through \autoref{eq:odds_decomp}, this can be achieved in two complementary ways:
\begin{itemize}
    \vspace*{-8px}
    \item \textbf{Strengthening visual grounding:} increasing the likelihood ratio in favor of absence, i.e., making $p_\theta(v'\mid z=0,x)$ relatively larger than $p_\theta(v'\mid z=1,x)$ by learning features that recognize missing objects / masked regions.
    \vspace*{-5px}
    \item \textbf{Weakening language priors:} reducing the prior odds $\frac{p_\theta(z=1\mid x)}{p_\theta(z=0\mid x)}$, i.e., learning that certain objects are \emph{not necessary} given $x$ and the remaining scene context.
    \vspace*{-5px}
\end{itemize}
Therefore, HIIs directly target the failure mode where language priors overwhelm visual evidence, by repeatedly presenting situations where the prior expectation is wrong but the visual signal is clear.

\vspace*{-3px}
\paragraph{Trade-off.}
Reducing overly strong language priors can sometimes lower performance when the prompt is under-specified and the model must rely on commonsense completion.
We discuss this capability--hallucination trade-off in~\autoref{sec::experiments}.

\section{Experiments on Qwen-Family}
\label{sub::exp_qwen}
\newcommand{\possmall}[1]{\textcolor{green!50!black}{\scalebox{0.85}{{+#1}}}}
\newcommand{\negsmall}[1]{\textcolor{red!70!black}{\scalebox{0.85}{{-#1}}}}

\begin{table*}[h]
\centering
\caption{\textbf{Experimental results on the Qwen-family models.}}
\label{tab::qwen} 
\resizebox{\textwidth}{!}{
\begin{tabular}{llllll|lll}
\toprule
\multirow{3}{*}{\textbf{Model}} & \multirow{3}{*}{\textbf{Method}} & \multicolumn{4}{c|}{\textbf{Hallucination benchmarks}} & \multicolumn{3}{c}{\textbf{General benchmarks}} \\ 
\cmidrule(lr){3-6} \cmidrule(lr){7-9}
 & & \multicolumn{2}{c}{\textbf{Object HalBench}} & \multicolumn{2}{c|}{\textbf{MOH}} & \textbf{TextVQA} & \textbf{ScienceQA} & \textbf{MM-Vet} \\
\cmidrule(lr){3-4} \cmidrule(lr){5-6} \cmidrule(lr){7-7} \cmidrule(lr){8-8} \cmidrule(lr){9-9}
 & & CHAIR\_s. $\downarrow$ & CHAIR\_i. $\downarrow$ & HR$^D$. $\downarrow$ & HR$^G$. $\downarrow$ & Acc. $\uparrow$ & Image Acc. $\uparrow$ & Overall $\uparrow$ \\
\midrule

\multirow{2}{*}{Qwen2-VL-2B-Instruct} 
 & baseline & 15.3 & 8.6 & 4.1 & 59.0 & 78.3 & 76.9 & 49.4 \\
 & \textbf{Ours} 
   & 2.9~\possmall{12.4} 
   & 1.7~\possmall{6.9} 
   & 4.7~\negsmall{0.6} 
   & 15.5~\possmall{43.5} 
   & 78.2~\negsmall{0.1} 
   & 76.9~\possmall{0.0}
   & 50.3~\possmall{0.9} \\
\midrule

\multirow{2}{*}{Qwen2-VL-7B-Instruct} 
 & baseline & 14.3 & 8.5 & 11.5 & 58.0 & 82.2 & 85.7 & 62.7 \\
 & \textbf{Ours} 
   & 4.2~\possmall{10.1} 
   & 2.5~\possmall{6.0} 
   & 9.0~\possmall{2.5} 
   & 20.0~\possmall{38.0} 
   & 82.3~\possmall{0.1} 
   & 86.7~\possmall{1.0} 
   & 62.7~\possmall{0.0}  \\
\midrule

\multirow{2}{*}{Qwen2.5-VL-7B-Instruct} 
 & baseline & 15.0 & 9.2 & 3.6 & 47.0 & 77.7 & 88.6 & 72.0 \\
 & \textbf{Ours} 
   & 2.8~\possmall{13.2} 
   & 1.6~\possmall{7.6} 
   & 2.2~\possmall{1.4} 
   & 17.4~\possmall{29.6} 
   & 78.4~\possmall{0.7} 
   & 88.5~\negsmall{0.1} 
   & 69.8~\negsmall{2.2} \\

\bottomrule
\end{tabular}
}
\end{table*}

To demonstrate the architectural robustness of our approach, we extend our evaluation to the latest Qwen2-VL and Qwen2.5-VL series. As shown in \autoref{tab::qwen}, the empirical results indicate that our method consistently mitigates hallucinations across both standard benchmarks and the specialized MOH benchmark. While certain metrics on general capability benchmarks exhibit marginal degradation, the overall performance remains highly competitive. This confirms that HII-DPO successfully suppresses scene-conditioned hallucinations without compromising the model's general understanding capability.

\section{Examples of the Synonym Dictionary for Entity Normalization.}
\label{sub::synonym_dict}
We built the synonym dictionary based on the list from~\cite{Lu_2018_CVPR,rohrbach2018object}. Here are some examples of the synonym dictionary:
\begin{description}[style=multiline, leftmargin=3cm, font=\bfseries, itemsep=2pt]
    \item[Person] \{girl, boy, man, woman, kid, child, chef, baker, people, rider, children, worker, sister, brother, \dots\}
    \item[Bird] \{ostrich, owl, seagull, goose, duck, parakeet, falcon, robin, pelican, waterfowl, heron, \dots\}
    \item[Dog] \{puppy, beagle, pup, chihuahua, schnauzer, dachshund, rottweiler, canine, pitbull, collie, \dots\}
    \item[Horse] \{colt, pony, racehorse, stallion, equine, mare, foal, palomino, mustang, clydesdale, bronco, \dots\}
    \item[Cow] \{cattle, oxen, ox, calf, holstein, heifer, buffalo, bull, zebu, bison\}
    \item[Boat] \{ship, liner, sailboat, motorboat, dinghy, powerboat, speedboat, canoe, skiff, yacht, kayak, vessel, \dots\}
    \item[Car] \{automobile, van, minivan, sedan, suv, hatchback, cab, jeep, coupe, taxicab, limo, taxi\}
    \item[Airplane] \{jetliner, plane, air plane, monoplane, aircraft, jet, airbus, biplane, seaplane\}
    \item[Train] \{locomotive, tramway, caboose\}
    \item[Traffic Light] \{street light, traffic signal, stop light, streetlight, stoplight\}
    \item[Sandwich] \{burger, sub, cheeseburger, hamburger\}
    \item[Cake] \{cheesecake, cupcake, shortcake, coffeecake, pancake\}
    \item[Couch] \{sofa, recliner, futon, loveseat, settee, chesterfield\}
    \item[Laptop] \{computer, notebook, netbook, lenovo, macbook, laptop computer\}
    \item[Phone] \{cell phone, mobile phone, cellphone, telephone, phon, smartphone, iPhone\}
\end{description}

\section{More Ablation Studies}
\label{sub::more_ablation}

\begin{table}[h]
\centering
\caption{\textbf{Ablation study on the response generation strategy.}}
\label{tab::sentence_by_sentence}
\resizebox{\columnwidth}{!}{\begin{tabular}{lcccccc}
\toprule
\multirow{2}{*}{\textbf{Method}} & \multicolumn{2}{c}{\textbf{Object HalBench}} & \multicolumn{3}{c}{\textbf{AMBER}} & \textbf{ScienceQA} \\ 
\cmidrule(lr){2-3} \cmidrule(lr){4-6} \cmidrule(lr){7-7}
 & CHAIR\_s. $\downarrow$ & CHAIR\_i. $\downarrow$ & CHAIR $\downarrow$ & Hal $\downarrow$ & Cog $\downarrow$ & Image Acc. $\uparrow$ \\ 
\midrule
Baseline & 52.7 & 27.9 & 8.4 & 35.5 & 4.0 & 66.8 \\
whole response & 12.5 & 4.6 & 3.6 & 16.9 & 1.6 & 67.5 \\
sentence-by-sentence & \textbf{4.0} & \textbf{2.5} & \textbf{2.7} & \textbf{13.5} & \textbf{0.9} & \textbf{68.6} \\
\bottomrule
\end{tabular}}
\end{table}

\paragraph{Sentence-by-sentence strategy.} We also investigate the effectiveness of the {sentence-by-sentence} strategy in dataset generation. As detailed in \autoref{tab::sentence_by_sentence}, we conduct a comparative analysis between the {whole-response} and {sentence-by-sentence} strategies. The results suggest that the latter achieves superior performance across both hallucination benchmarks and general VQA tasks. We hypothesize that this improvement stems from two primary factors \textbf{(1) Granularity of Reward Signals:} The whole-response strategy often yields rejected responses that conflate factual and hallucinated information simultaneously \cite{gu2025maskdpogeneralizablefinegrainedfactuality}, which can lead to corrupted reward signals during preference alignment. Conversely, the sentence-by-sentence strategy constructs preference pairs at the individual sentence level, ensuring fine-grained and more precise optimization signals. \textbf{(2) Enhanced Data Diversity:} While the whole-response strategy typically generates only a single preference pair per image, the sentence-by-sentence approach enables the extraction of multiple preference pairs from a single image-response interaction, significantly increasing the diversity of the training data.

\begin{table}[h]
\centering
\caption{\textbf{Ablation study on different prompt types ($N$).}}
\label{tab::prompts_types}
\resizebox{\columnwidth}{!}{\begin{tabular}{lccccccc}
\toprule
\multirow{2}{*}{\textbf{N}} & \multicolumn{2}{c}{\textbf{Object HalBench}} & \multicolumn{3}{c}{\textbf{AMBER}} & \textbf{ScienceQA} & \textbf{MM-Vet} \\ 
\cmidrule(lr){2-3} \cmidrule(lr){4-6} \cmidrule(lr){7-7} \cmidrule(lr){8-8}
 & CHAIR\_s. $\downarrow$ & CHAIR\_i. $\downarrow$ & CHAIR $\downarrow$ & Hal $\downarrow$ & Cog $\downarrow$ & Image Acc. $\uparrow$ & Overall $\uparrow$ \\ 
\midrule
Baseline & 52.7 & 27.9 & 8.4 & 35.5 & 4.0 & 66.8 & 31.0 \\
2 & {7.2} & {4.0} & {3.2} & {14.5} & {1.2} & {67.9} & {32.9} \\
3  & {4.0} & {2.5} & {2.7} & {13.5} & \textbf{0.9} & \textbf{68.6} & \textbf{33.8} \\
4  & \textbf{2.9} & \textbf{2.3} & \textbf{2.5} & \textbf{11.9} & \textbf{0.9} & {67.5} & {33.2} \\
\bottomrule
\end{tabular}}
\end{table}

\paragraph{Prompts in dataset generation.} To ensure diversity in dataset generation, we utilize multiple instructions for each HII when generating preference datasets. Specifically, we randomly select $N$ prompts from the following collection:

\begin{itemize}[noitemsep, topsep=0pt]
    \item \textit{``What is this photo about? Please answer in great detail.''}
    \item \textit{``Describe this image in detail.''}
    \item \textit{``Provide a thorough description of the given picture.''}
    \item \textit{``Please provide a detailed description of the picture.''}
    \item \textit{``Take a look at this image and describe what you notice.''}
    \item \textit{``Could you describe the contents of this image for me?''}
\end{itemize}

We vary the value of $N$ and present the corresponding experimental results in \autoref{tab::prompts_types}. We observe that using only one prompt per HII results in suboptimal alignment due to a lack of diversity. Conversely, although a larger $N$ (i.e., $N > 3$) further reduces hallucinations, it tends to cause overfitting, which undermines the model's underlying knowledge and general VQA performance. Overall, $N=3$ strikes the ideal balance, effectively mitigating hallucinations while maintaining robust general-purpose understanding.

\begin{table*}[t]
\centering
\caption{Ablation study on model-specific HII filtering.}
\label{tab::randomly_mask} 
\resizebox{\textwidth}{!}{
\begin{tabular}{lccccccc|cccc}
\toprule
\multirow{3}{*}{\textbf{Method}} 
& \multicolumn{7}{c|}{\textbf{Hallucination benchmarks}} 
& \multicolumn{4}{c}{\textbf{General benchmarks}} \\ 
\cmidrule(lr){2-8} \cmidrule(lr){9-12}
& \multicolumn{2}{c}{\textbf{Object HalBench}} 
& \multicolumn{3}{c}{\textbf{AMBER}} 
& \multicolumn{2}{c|}{\textbf{MOH}}
& \textbf{VQAv2} & \textbf{TextVQA} & \textbf{ScienceQA} & \textbf{MM-Vet} \\
\cmidrule(lr){2-3} \cmidrule(lr){4-6} \cmidrule(lr){7-8}
\cmidrule(lr){9-9} \cmidrule(lr){10-10} \cmidrule(lr){11-11} \cmidrule(lr){12-12}
& CHAIR\_s. $\downarrow$ & CHAIR\_i. $\downarrow$ 
& CHAIR $\downarrow$ & Hal. $\downarrow$ & Cog. $\downarrow$ 
& HR$^D$. $\downarrow$ & HR$^G$. $\downarrow$ 
& Acc. $\uparrow$ & Acc. $\uparrow$ & Image Acc. $\uparrow$ & Overall $\uparrow$ \\
\midrule
baseline 
& 52.7 & 28.0 & 8.4 & 35.5 & 4.0 
& 52.2 & 68.6 
& \textbf{78.5} & \textbf{58.2} & 66.8 & 31.0 \\
random masking 
& 9.3 & 3.6 & 3.2 & 15.8 & 1.6 
& 40.3 & 37.2 
& 78.0 & \textbf{58.2} & 68.4 & 32.8 \\
model-specific HII 
& \textbf{4.0} & \textbf{2.5} & \textbf{2.7} & \textbf{13.5} & \textbf{0.9} 
& \textbf{28.1} & \textbf{18.8} 
& 77.8 & \textbf{58.2} & \textbf{68.6} & \textbf{33.8} \\
\bottomrule
\end{tabular}
}
\end{table*}

\textbf{Random masking vs. model-specific HIIs.}
We additionally perform an ablation by replacing model-specific HII filtering with {random} object masking, while keeping {all other components identical} (the same detector, masking procedure, and preference-pair construction). \autoref{tab::randomly_mask} shows that random masking already alleviates hallucination, suggesting that high-quality counterfactual preference pairs (facilitated by GroundingDINO for accurate object localization and consistent grounding) provide a generally useful alignment signal.
Nevertheless, random masks are not semantically tied to the underlying scenarios. Thus, the removed objects often do not correspond to the objects that the model would hallucinate due to strong scene-object co-occurrence priors.
As a result, the preference pairs are less informative for correcting the model’s dominant failure mode (scene-conditioned hallucination), leading to weaker improvements than our {model-specific} HII filtering, which explicitly focuses on the scene-conditioned hallucination pattern.

\section{Training Details}

\subsection{Dataset}
\paragraph{MSCOCO.}MSCOCO ~\cite{lin2015microsoftcococommonobjects} is a large-scale object detection, segmentation, and captioning dataset. It comprises over 328,000 images, covering 80 object categories with more than 2.5 million labeled instances. For vision-language tasks, each image is paired with at least five human-written captions, ensuring a standard benchmark for evaluating image captioning and cross-modal retrieval models.

\paragraph{Visual Genome.}Visual Genome~\cite{krishna2017visual} is a densely annotated visual dataset that focuses on detailed scene semantics beyond object presence. It provides fine-grained annotations including objects, attributes, relationships, region descriptions, and scene graphs for each image. Compared to MSCOCO~\cite{lin2015microsoftcococommonobjects}, Visual Genome offers a more exhaustive coverage of object categories and relational structures, enabling more precise analysis of object interactions and compositional semantics. These characteristics make Visual Genome particularly valuable for studying scene-conditioned hallucination patterns and for synthesizing challenging counterfactual or hallucination-inducing samples where subtle visual cues and relational dependencies are critical.

\begin{table}[htbp]
\centering
\caption{Training hyperparameters for LLaVA-v1.5 models. Note: Weight decay is 0. Memory optimization uses Zero-2~\cite{rajbhandari2020zero}.}
\label{tab::hyperparameters}
\resizebox{\textwidth}{!}{%
\begin{tabular}{l|cccccccccc}
\toprule
\textbf{Base Model} & \textbf{lr} & \textbf{LoRA $r$} & \textbf{LoRA $\alpha$} & \textbf{LoRA $\beta$} & \textbf{Proj. LR} & \textbf{Scheduler} & \textbf{Optim} & \textbf{Length} & \textbf{Epochs} & \textbf{Global Batch} \\
\midrule
LLaVA-v1.5-7B  & 2e-6 & 128 & 256 & 0.1 & 0 & Cosine & AdamW & 2048 & 1 & 64 \\
LLaVA-v1.5-13B & 3e-6 & 128 & 256 & 0.1 & 0 & Cosine & AdamW & 2048 & 1 & 64 \\
\bottomrule
\end{tabular}%
}
\end{table}
\subsection{Training Setup}
In order to ensure a fair comparison with existing methods \cite{peng2025mitigatingobjecthallucinationssentencelevel, xing2025realignaligningvisionlanguage, xiao2025detectingmitigatinghallucinationlarge, wang2024mdpoconditionalpreferenceoptimization, he2024topic, zhao2024hallucinationsenhancinglvlmshallucinationaware, zhou2024aligningmodalitiesvisionlarge, yu2025rlaif}, we employ LLaVA-v1.5-7B and LLaVA-v1.5-13B \cite{liu2023visualinstructiontuning} as our backbone models. We conduct preference alignment using the proposed HII-DPO framework. For the training configuration, we utilize LoRA fine-tuning with a rank of $r = 128$ and an alpha of $\alpha = 256$. Both the 7B and 13B models are fine-tuned for 1 epoch, with learning rates set to $2 \times 10^{-6}$ and $3 \times 10^{-6}$, respectively. More details about hyperparameters are presented in~\autoref{tab::hyperparameters}.


\section{HII examples}
\label{sec::case_MOH}
\begin{figure}[htbp]
    \centering
    \includegraphics[trim=0em 16em 0em 10em, clip, width=1\textwidth]{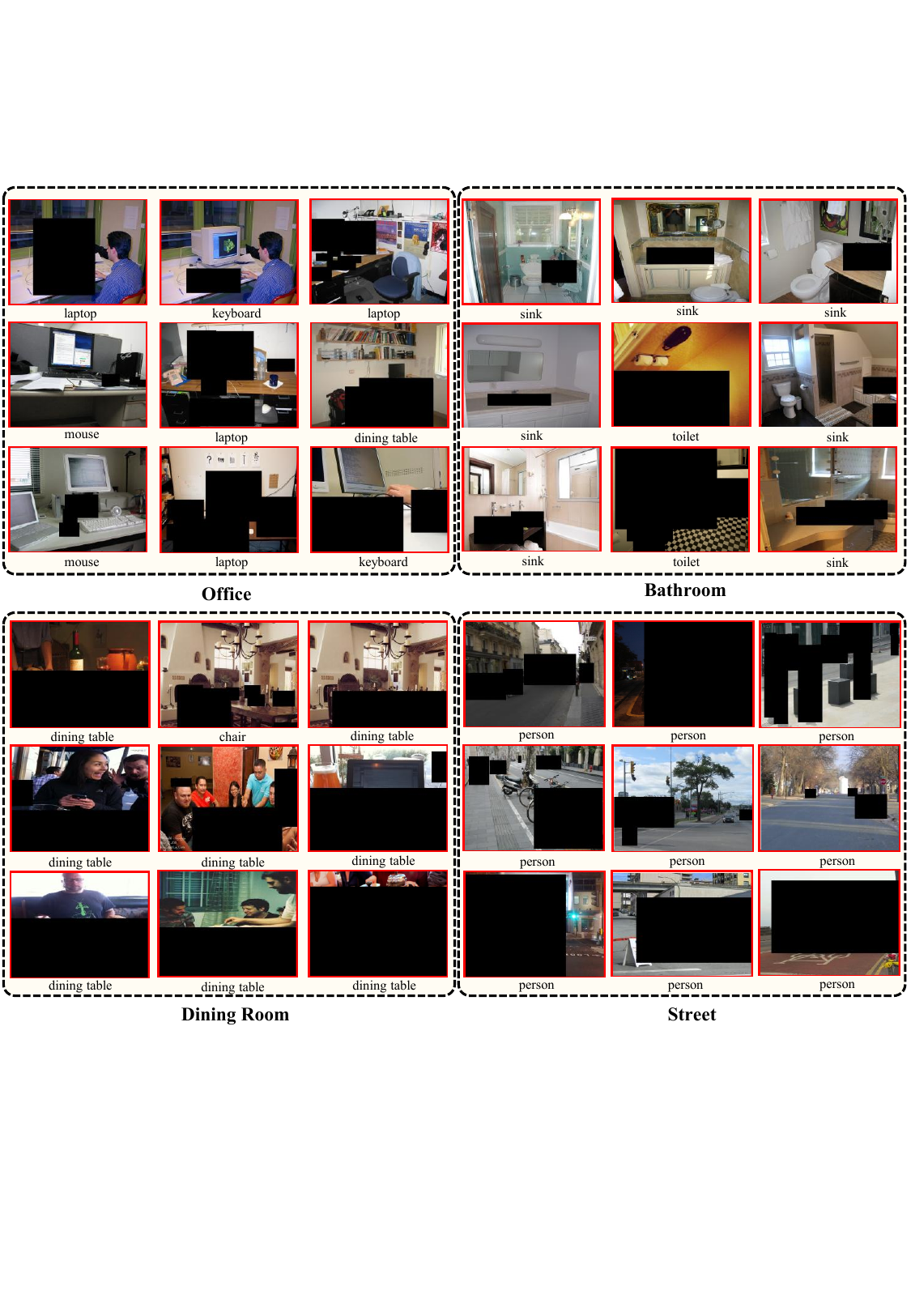} 
    \caption{HII examples in \textbf{Office, Bathroom, Dining Room and Street} scenarios.}
    \label{fig::HII-cases1}
\end{figure}

\begin{figure}[htbp]
    \centering
    \resizebox{0.90\textwidth}{0.90\height}{\includegraphics[trim=0em 0em 0em 0em, clip, width=1\textwidth]{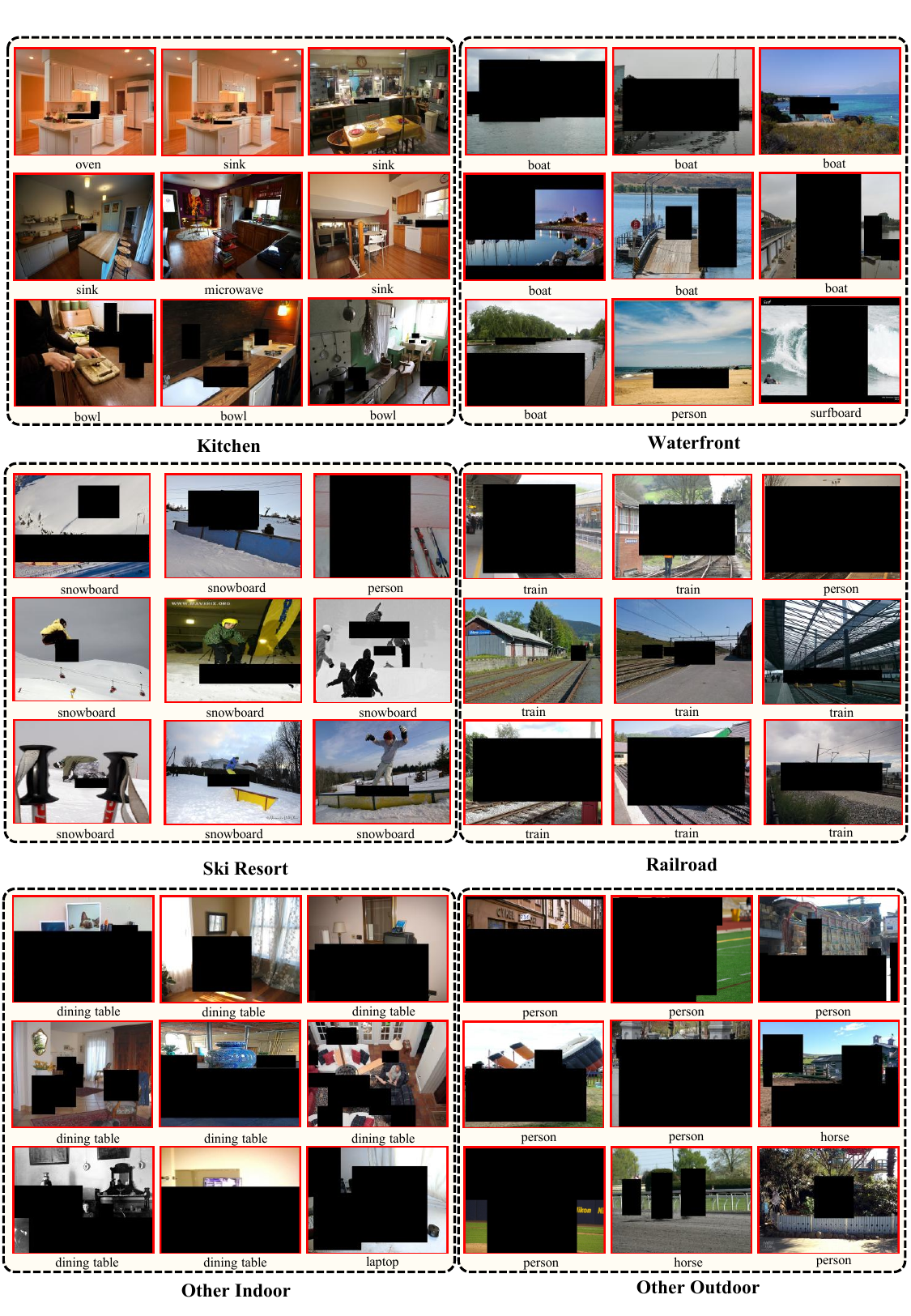}}
    \caption{HII examples in \textbf{Kitchen, Waterfront, Ski Resort, Railroad, Other Indoor and Other Outdoor} scenarios.}
    \label{fig::HII-cases2}
\end{figure}

In this section, we present more HII examples to illustrate the scene-conditioned hallucination patterns in VLMs.
Following our HII synthesis pipeline~(\autoref{fig::HIIs_pipeline}), we generate {model-specific} HIIs using VLMs of different scales and architectures.
We then take the intersection of these model-specific HII sets to obtain the {hardest} counterfactual images---those that consistently trigger the same hallucinated object across models---and use them to construct MOH for rigorous evaluation of scene-conditioned hallucinations.
Finally, we provide qualitative examples in~\autoref{fig::HII-cases1} and \autoref{fig::HII-cases2}, which highlight how scene context induces systematic object hallucinations despite the absence of visual evidence.


\section{Additional Case Studies}
\label{sec::add_cases}
\begin{figure}[htbp]
    \centering
    \resizebox{0.90\textwidth}{0.90\height}{\includegraphics[trim=0em 0em 0em 0em, clip, width=1\textwidth]{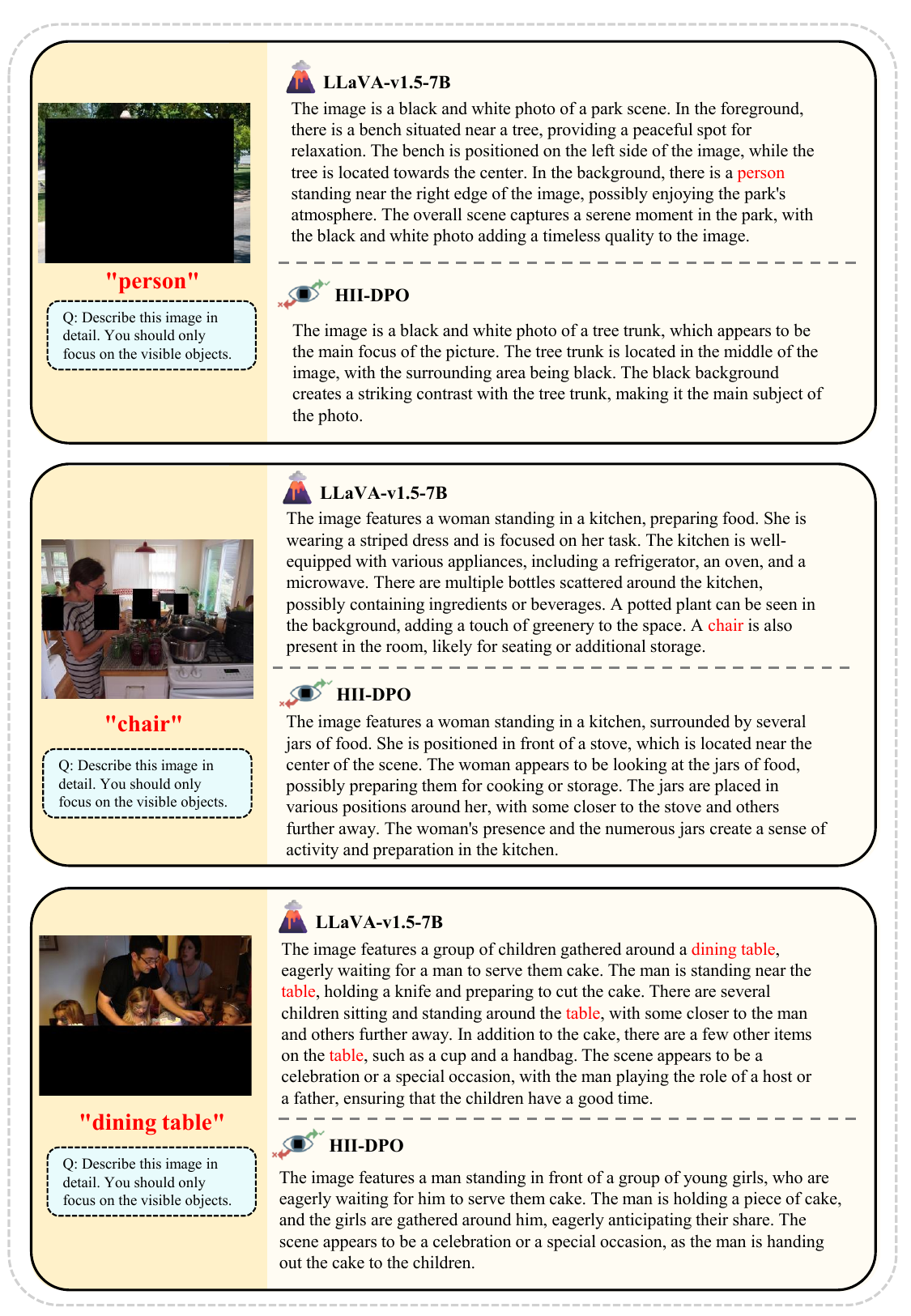}} 
    \caption{A qualitative comparison of Detailed Description Generation (DDG) between the base LLaVA-v1.5-7B and HII-DPO.}
    \label{fig::more-case1}
\end{figure}

\begin{figure}[htbp]
    \centering
    \resizebox{0.90\textwidth}{0.90\height}{\includegraphics[trim=0em 0em 0em 0em, clip, width=1\textwidth]{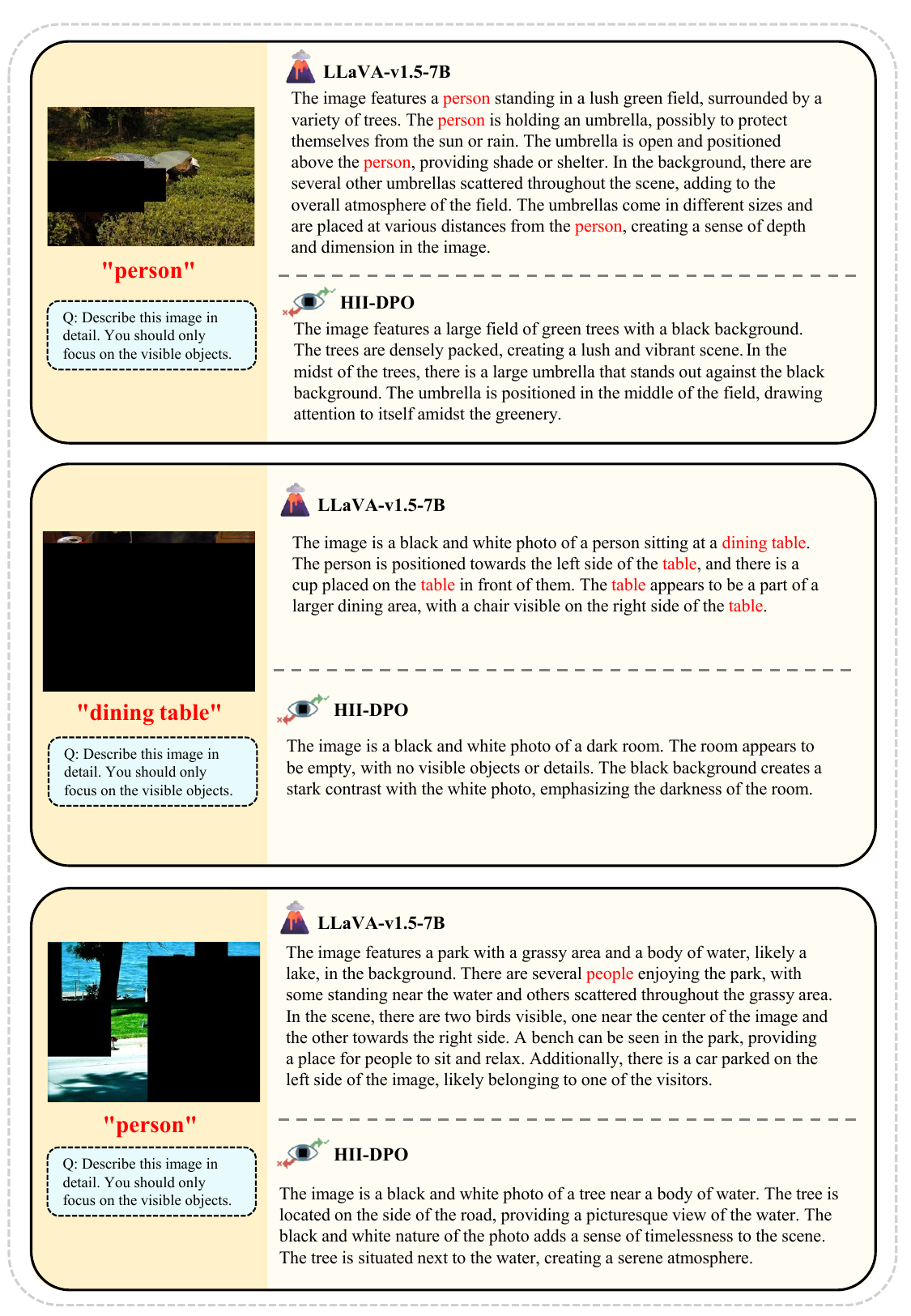}}
    \caption{Another qualitative comparison of Detailed Description Generation (DDG) between the base LLaVA-v1.5-7B and HII-DPO.}
    \label{fig::more-case2}
\end{figure}

To further substantiate the efficacy of our approach in mitigating the scene-conditioned hallucination pattern, we provide additional qualitative examples in \autoref{fig::more-case1} and \autoref{fig::more-case2}. We compare the responses generated by the base model LLaVA-v1.5 and HII-DPO for the same HII. Given a Hallucination-Inducing Image (HII) where the specific object is removed, the base model repeatedly hallucinates the masked object and related items due to rooted scene-object co-occurrence hallucination pattern. Conversely, HII-DPO accurately describes the scene without relying on linguistic heuristics.

\end{document}